\documentclass[journal]{IEEEtran}
\usepackage{times}

\usepackage[numbers]{natbib}
\usepackage{multicol}
\usepackage[bookmarks=true]{hyperref}

\usepackage{hyperref}
\usepackage{url}
\usepackage{graphicx}
\usepackage{float}

\usepackage{algorithm}
\usepackage{amsmath} 
\usepackage{amssymb}  
\usepackage{dsfont} 
\usepackage[noend]{algpseudocode}
\usepackage[T1]{fontenc}
\usepackage[utf8]{inputenc}

\usepackage{booktabs}
\usepackage{tabularx}
\usepackage{array}
\usepackage{xcolor}

\usepackage{subfigure}

\begin{document}

\title{Behavior-Constrained Reinforcement Learning with Receding-Horizon Credit Assignment for High-Performance Control}

\author{Siwei Ju, Jan Tauberschmidt, Oleg Arenz, Peter van Vliet, Jan Peters%
\thanks{Siwei Ju, Oleg Arenz, and Jan Peters are with the Intelligent Autonomous Systems Lab, Department of Computer Science, TU Darmstadt, Germany.}%
\thanks{Siwei Ju and Peter van Vliet are with Porsche AG, Stuttgart, Germany.}%
\thanks{Jan Tauberschmidt is with the German Research Center for Artificial Intelligence (DFKI), Germany.}%
}
\maketitle

\begin{abstract}

Learning high-performance control policies that remain consistent with expert behavior is a fundamental challenge in robotics. Reinforcement learning can discover high-performing strategies but often departs from desirable human behavior, whereas imitation learning is limited by demonstration quality and struggles to improve beyond expert data. This challenge is particularly pronounced in high-performance dynamic systems, where desirable behavior is expressed over trajectories and the consequences of suboptimal decisions often emerge only after a delay.
We propose a behavior-constrained reinforcement learning framework that improves beyond demonstrations while explicitly controlling deviation from expert behavior. Because expert-consistent behavior in dynamic control is inherently trajectory-level, we introduce a receding-horizon predictive mechanism that models short-term future trajectories and provides look-ahead rewards during training. To account for the natural variability of human behavior under disturbances and changing conditions, we further condition the policy on reference trajectories, allowing it to represent a distribution of expert-consistent behaviors rather than a single deterministic target.
Empirically, we evaluate the approach in high-fidelity race car simulation using data from professional drivers, a domain characterized by extreme dynamics and narrow performance margins. The learned policies achieve competitive lap times while maintaining close alignment with expert driving behavior, outperforming baseline methods in both performance and imitation quality. Beyond standard benchmarks, we conduct human-grounded evaluation in a driver-in-the-loop simulator and show that the learned policies reproduce setup-dependent driving characteristics consistent with the feedback of top-class professional race drivers.
These results demonstrate that our method enables learning high-performance control policies that are both optimal and behavior-consistent, and can serve as reliable surrogates for human decision-making in complex control systems.
\end{abstract}

\section{Introduction}

Many robotic control problems require more than maximizing task performance alone. In practical applications, controllers must often remain consistent with expert or human-preferred behavior while operating near the limits of system capability. This is particularly important in high-performance dynamic control, where small differences in control strategy can strongly affect both performance and system behavior. Learning such policies is challenging: purely reward-driven optimization may discover highly effective but undesirable strategies, whereas strict imitation of demonstrations can limit adaptation, robustness, and improvement beyond expert data. This tension is especially pronounced in dynamic systems where both task success and the manner in which it is achieved are critical.

Professional motorsports provide a compelling instance of this challenge. Race car driving requires operating a nonlinear system near its physical limits, where small differences in control strategy can have large consequences for both performance and stability. The task is further complicated by delayed effects of control decisions, narrow performance margins, strong coupling between driver behavior and vehicle dynamics, and the existence of multiple valid but stylistically distinct driving strategies. These properties make race driving not only a demanding benchmark for learned control, but also a practically important domain in which expert-consistent behavior is essential rather than optional.

A particularly important application in this setting is \emph{virtual setup testing}, where engineers iteratively adjust vehicle \emph{setup}, including suspension, aerodynamics, and powertrain parameters, to optimize the coupled driver--vehicle system under changing conditions. Traditionally, this process has relied on control-theoretic methods grounded in vehicle dynamics models, such as linear quadratic programming (LQR) and model predictive control (MPC), to evaluate candidate setups and design track-specific controllers~\cite{optimal_control, mpc}. These approaches offer transparency and strong guarantees when models are accurate, but require extensive modeling, identification, and calibration, and can become less effective near the handling limits where nonlinearities dominate and driver skill becomes decisive. In contrast, data-driven policies learned from simulation and demonstration offer the possibility of capturing complex limit behavior while enabling rapid and low-cost virtual experimentation~\cite{lockel_adaptive_2023,ju_digital_twin_2023}.

\begin{figure}
    \centering
    \includegraphics[width=\columnwidth]{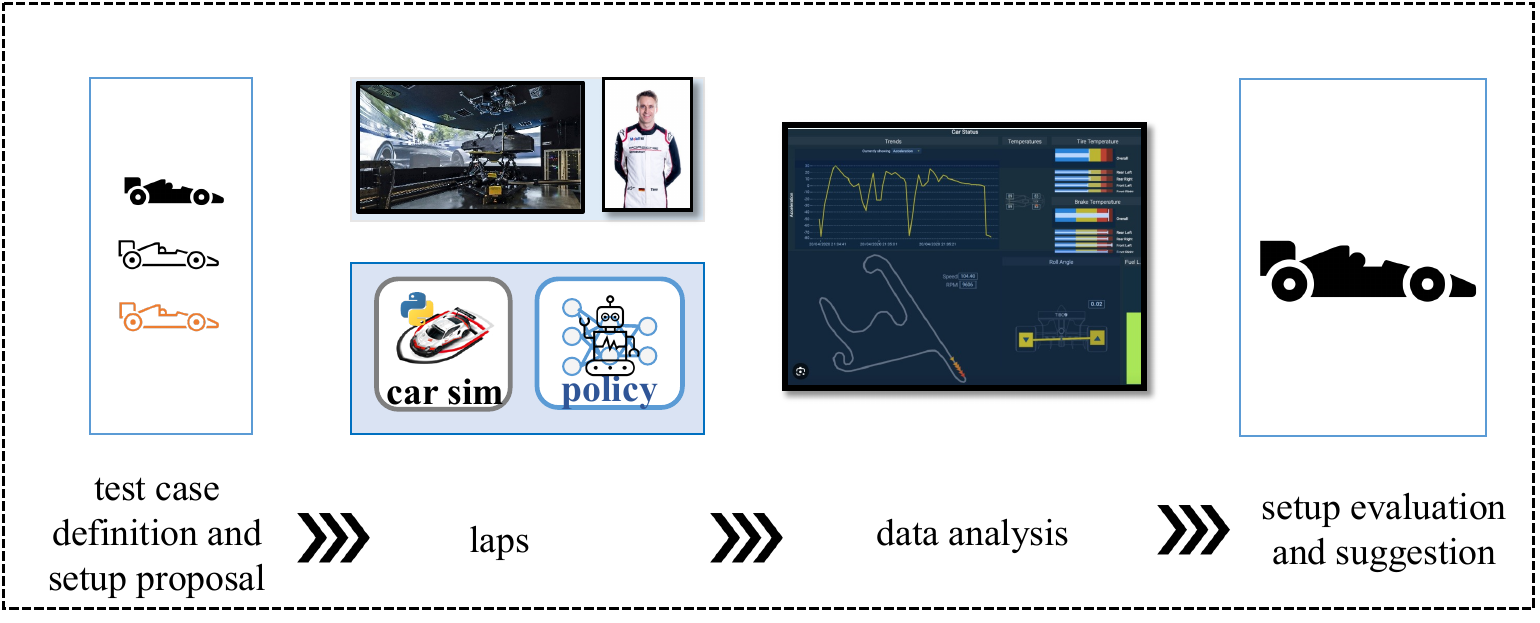}
    \vspace{-20pt}
    \caption{Application workflow of the proposed method in professional motorsports. A learned policy acts as a behavior-consistent digital driver model for virtual setup evaluation in a driver-in-the-loop simulation environment.}
    \label{fig:application}
\end{figure}

Recent successes in reinforcement learning (RL) have shown that learned policies can achieve, and in some settings surpass, elite human performance in complex control domains including autonomous driving and racing~\cite{PPO, ju_digital_twin_2023, sonyAI_2022}. These results highlight the strength of RL for optimizing well-defined task objectives through trial-and-error interaction. However, in applications such as virtual setup testing, raw task performance alone is insufficient. A policy that achieves a fast lap by exploiting simulator-specific strategies or adopting unrealistic control patterns is of limited practical value if it no longer reflects how a professional driver would interact with the vehicle. More generally, reward-maximizing policies may exploit proxy objectives, converge to brittle local optima, or produce behaviors that are difficult to interpret and trust in human-referenced evaluation settings.

Imitation learning (IL) offers a complementary route by learning directly from expert demonstrations. Methods such as Behavioral Cloning and GAIL can efficiently recover expert-like behavior and are commonly used to initialize or accelerate policy learning~\cite{deepMimic, benchmark_IL_racing, lockel_adaptive_2023}. However, IL is fundamentally limited by the quality and coverage of the demonstration dataset and remains vulnerable to covariate shift and compounding errors under distribution shift~\cite{covariate_shift}. In race car driving, this limitation is particularly pronounced: even modest changes in vehicle setup can alter the dynamics enough to expose generalization gaps despite extensive expert data~\cite{ju_digital_twin_2023, lockel_adaptive_2023, sonyAI_2022}. This raises a central question for high-performance control: how can a policy improve beyond demonstrations while remaining behaviorally consistent with expert driving?

A further challenge is that expert-consistent behavior in dynamic control is inherently \emph{trajectory-level}. In racing, desirable behavior is not determined only by instantaneous actions or local tracking errors, but by how a sequence of decisions unfolds over time. For example, a slightly excessive corner-entry speed may only manifest as understeer or loss of track position several seconds later. Temporal credit assignment is therefore crucial not only for optimizing performance, but also for learning behavior that remains consistent with expert strategies. While predictive models can in principle support such reasoning~\cite{CAP_survey}, accurate long-horizon modeling near the limits of adhesion remains difficult. At the same time, human driving behavior is not deterministic: natural variability, adaptation to disturbances, and setup-dependent strategy changes lead to a \emph{distribution} of valid expert trajectories rather than a single nominal reference.

In this work, we address these challenges with a \emph{behavior-constrained reinforcement learning} framework for high-performance control. We formulate policy improvement as a constrained optimization problem that maximizes task performance while explicitly limiting deviation from expert behavior. To better capture the short-horizon trajectory structure underlying expert driving, we introduce a lightweight receding-horizon trajectory predictor based on probabilistic B\'ezier curves, which provides both an auxiliary learning signal and look-ahead rewards for improved temporal credit assignment. To account for the natural variability of human behavior, we further condition the policy on reference trajectories, enabling it to represent a distribution of expert-consistent behaviors rather than collapsing demonstrations to a single deterministic target. Finally, we combine this formulation with targeted data augmentation and supervised pretraining to improve robustness and training efficiency.

We evaluate the proposed method in high-fidelity race car simulation using demonstration data collected from professional drivers on a driver-in-the-loop simulator. This setting provides a particularly demanding benchmark due to extreme vehicle dynamics, narrow performance margins, and the practical importance of human-consistent behavior for setup evaluation. Beyond standard performance and imitation metrics, we perform human-grounded evaluation across multiple vehicle setups and show that the learned policies reproduce setup-dependent driving characteristics consistent with expert driver feedback. This suggests that learned policies can serve not only as high-performance controllers, but also as behavior-consistent digital surrogates for human-in-the-loop system evaluation.

Our main contributions are:
\begin{itemize}
    \item \textbf{Behavior-constrained policy optimization.} We formulate high-performance driving as a constrained reinforcement learning problem that maximizes task performance while enforcing a lower bound on trajectory consistency with expert demonstrations.
    
    \item \textbf{Receding-horizon credit assignment.} We introduce a lightweight short-horizon trajectory predictor based on probabilistic B\'ezier curves and use it to provide look-ahead rewards that improve temporal credit assignment in dynamic control.
    
    \item \textbf{Trajectory-conditioned expert behavior modeling.} We condition the policy on reference trajectories to represent the natural variability of expert driving under disturbances and changing conditions, and combine this with targeted data augmentation for improved robustness.
    
    \item \textbf{Human-grounded validation in motorsports.} We demonstrate the method in high-fidelity race car simulation and driver-in-the-loop evaluation, showing strong performance, improved imitation quality, and setup-dependent behavior consistent with feedback from professional race drivers.
\end{itemize}
\section{Related Work}
\label{sec:related_work}
The multi-objective nature of our problem cuts across several research lines. We first review imitation learning (IL) and hybrid RL–IL formulations as general methodologies for combining objectives. We then discuss techniques that strengthen learning signals and credit assignment, including predictive modeling and auxiliary tasks. Finally, we cover applications to autonomous racing, where the goal extends beyond achieving pace to reproducing professional driving style.

\paragraph*{Imitation Learning and Reinforcement Learning}
Learning from demonstrations can be framed as minimizing a divergence between expert and learner trajectory distributions, with different divergences inducing common IL algorithms \cite{il_divergence_minimization, il_divergence_minimization_2}. Behavioral Cloning (BC) \cite{BC} is a simple supervised approach but suffers from covariate shift and compounding errors \cite{covariate_shift}. Dataset Aggregation (DAgger) \cite{DAgger} mitigates these issues by iteratively collecting expert corrections. Adversarial approaches such as GAIL \cite{GAIL} let the policy interact with the environment while a discriminator provides a learned reward; variants tailor this idea to specific domains, including driving and multi-style imitation \cite{modeling_driving_AIL, MultiGAIL, BeTAIL}. While IL can reach expert-like performance in favorable settings \cite{deepMimic, benchmark_IL_racing}, robustness hinges on demonstration quality and coverage, and transfer to new conditions remains difficult.

Combining performance objectives with style imitation is commonly realized by adding imitation terms to RL losses. DeepMimic \cite{deepMimic} augments extrinsic rewards with point-wise tracking of reference motions, whereas AMP \cite{amp} replaces hand-crafted distances with adversarial rewards to match style \cite{amp}. More broadly, multi-objective RL addresses conflicting goals via scalarization with utility functions \cite{multi_objective_survey, multi_objective_survey_2, mo_utility_survey}; dynamic utilities adapt weights during training \cite{dynamic_utility}. In contrast to using demonstrations purely to initialize or accelerate exploration, we explicitly treat demonstrations as an additional objective and control the performance–style trade-off through a constraint on imitation error.

\paragraph*{Trajectory Prediction}
Model-based learning and representation learning via prediction can improve credit assignment and sample efficiency. Latent world models \cite{world_models} and Dreamer-style agents \cite{dreamer} demonstrate the benefits of learning dynamics and rewards. Self-prediction objectives enhance representations by forecasting latent states over multiple steps \cite{self_prediction, understanding_self_predictive, self_predictive_efficient, when_self_predict}. Complementary to full dynamics modeling, auxiliary tasks align latent features with task-relevant signals and can substantially improve PPO and related methods \cite{aux_pointgoal, PPO}. Our approach follows this direction with a lightweight auxiliary predictor to aid near-horizon credit assignment.

\paragraph*{Applications to Racing}
Autonomous racing has advanced rapidly \cite{autonomous_on_edge}. Superhuman performance has been demonstrated in realistic racing games \cite{sonyAI_2022}, though matching human race-driver style remains challenging. Classical path-tracking relies on geometric controllers such as Pure Pursuit and Stanley \cite{pure_pursuit, stanley} and on MPC for predictive control \cite{mpc_survey}, with Bézier representations appearing in two-stage pipelines that first predict a curve and then track it \cite{deepracing}. RL provides an alternative that can either complement classical controllers or directly control the vehicle, with evidence favoring end-to-end RL for path tracking \cite{rl_path_tracking}. IL is frequently used to bootstrap RL \cite{benchmark_IL_racing, ju_digital_twin_2023} or in interactive settings \cite{mega_dagger}. Beyond raw performance, recent efforts target style: \cite{robust_player_imitation} combines progress rewards with BC in games; \cite{driving_fuzzy} introduces an action-similarity evaluator to transfer style to new tracks; \cite{lockel_identification_2022} classifies professional styles and proposes ProMoD, a BC-based framework that reproduces style differences but inherits BC’s limitations. Context conditioning on reference lines improves variability and realism in virtual setup testing \cite{ju_digital_twin_2023}.

\paragraph*{Style-Biased Reinforcement Learning}
Balancing task performance with stylistic fidelity has been studied in robot locomotion as \emph{Style-Biased Reinforcement Learning (SBRL)} \cite{sbrl}\footnote{This paper is currently under double-blind review, and we have attached the manuscript as additional material.}, often in tandem with motion-tracking objectives such as \cite{deepMimic} and adversarial priors \cite{amp}. Our work extends this paradigm to motorsport vehicle control. Relative to prior SBRL, we tailor the formulation to racing with contextual conditioning on target trajectories, strengthen imitation signals via probabilistic movement primitives (ProMP) \cite{promp} and targeted augmentation, and adopt domain-informed state, action, and reward designs. We provide quantitative comparisons to strong baselines and demonstrate a virtual setup-testing pipeline validated with driver-in-the-loop studies.

\section{Methodology}
In this paper, we aim to develop a human-like and competitive race car driving policy tailored for high-fidelity race simulations. To achieve this objective, the policy must effectively manage the complex and highly dynamic racing environment while achieving optimal performance and imitating human driving styles. To address this challenge, we propose a style-biased contextual reinforcement learning framework capable of handling both demonstrated and undemonstrated scenarios.
While the method is inspired by our problem of racing, we will first introduce it as a general framework.

\subsection{Problem Formulation}
Our proposed method aims to imitate human behavior including its inherent stochasticity. To this end, we leverage a contextual formulation that allows us to imitate an entire distribution of expert behavior.
Following the general definition of contextual reinforcement learning \cite{contextual_rl_2015}, we assume that the context $c$, representing full expert demonstrations of specific parts of expert behavior, is distributed according to some probability measure $\mathcal{C}$.
Since we want to imitate expert behavior while optimizing performance with respect to a predefined goal, the reward consists of potentially conflicting \textit{performance reward} $r^p$ for performance and \textit{style imitation reward} $r^s$:
\begin{equation}
\label{eq: sb-reward}
    r(s_t, a_t, c) = r^p(s_t, a_t, c) + \alpha^{s}r^s(s_t, a_t, c).
\end{equation}
The dependence of the reward on the context $c$ allows us to incentivize imitation of specific expert behaviors. We elaborate on each component in the reward later.

For performance-critical domains such as race car driving, it is essential that the agent achieves optimal performance even in environments not seen during training. This requires a careful balance between imitation fidelity and performance optimization, which is difficult to tune manually, as the optimal trade-off varies across environments (domain shift) and might evolve throughout training. Improperly chosen parameters can not only degrade performance but also destabilize learning.
To address this, we adopt the approach proposed in \cite{sbrl} and formulate the problem as a constrained optimization task, in which the goal is to maximize performance while enforcing a style imitation constraint denoted by $\hat{R}$:
\begin{equation}
    \max_\pi \mathbb{E}_\mathcal{C} \mathbb{E}_\pi\left[ \sum_{t=0}^T r(s_t, a_t, c)\right],\ \text{ s.t.}\ \mathbb{E}_\pi[r^s_t] \ge\hat R\ \  \forall t
\end{equation}

As introduced in \cite{sbrl}, the trade-off coefficient $\alpha^s$ in Eq.~\ref{eq: sb-reward} is interpreted as the Lagrangian multiplier for the original optimization problem and then updated along with the policy.

The loss function for $\alpha^s$ is written as 
\begin{equation}
\mathcal{L}_{\alpha^s} = \alpha^s\left( \hat{R} - \mathbb{E}_\pi[r^s] \right)
\label{eq:alpha}
\end{equation}

\subsection{Short-term trajectory representation and prediction}
In \cite{sbrl}, a lightweight trajectory predictor is trained as an auxiliary task to strengthen the model’s understanding of short-term consequences. We find such near-horizon awareness equally important for driving. 

For racing, the driving line is a result of all driver inputs and track conditions. Therefore, we propose to predict the short-term trajectory in the form of the coordinates on the driving line as the task-critical states for prediction, where all collected trajectories are transformed into the car's local coordinate system. 

A key challenge in this setting is the prediction horizon: it must be sufficiently long to capture context relevant for decision-making, yet extending it increases prediction complexity and dimensionality.
To address this trade-off, we introduce a compact but informative trajectory representation based on \emph{probabilistic Bézier curves}  \cite{bezier_curve}, which reduces the number of parameters the predictor must output while preserving key geometric structure.

Following \cite{sbrl}, let $\psi$ denote the predictor of desired future trajectories. We model these futures as a sequence of $H$ parametric distributions with parameters in the task-critical state space $\mathcal{D}$. Rather than predicting parameters for all $H$ future points directly, $\psi$ outputs parameters for $M{+}1$ control points, which are then interpolated to obtain the $H$-step trajectory (i.e., $\mathcal{D}^{M{+}1}\!\to\!\mathcal{D}^H$ via Bézier interpolation).

A probabilistic Bézier curve is determined by control points $(P_0,\dots,P_M)$, where each control point is distributed as ${P_i \sim \mathcal{N}(\mu_i,\Sigma_i)}$ with ${(\mu_i,\Sigma_i)\!\in\!\mathcal{D}}$. We take $P_0$ to be the current vehicle state, fix it as the origin of the local curve frame, and do not learn its parameters. For a phase variable $t\in[0,1]$, the mean and covariance along the curve are
\[
\mu_\psi(t)=\sum_{i=0}^M b_{i,M}(t)\,\mu_i,
\qquad
\Sigma_\psi(t)=\sum_{i=0}^M \bigl(b_{i,M}(t)\bigr)^2 \Sigma_i,
\]
where $b_{i,M}(t)=\binom{M}{i}(1-t)^{M-i}t^i$ are the Bernstein polynomials. To produce distributions for $H$ points $\{\psi_i\}_{i=1}^H$, the network first predicts $\{(\mu_i,\Sigma_i)\}_{i=0}^M$, and we then evaluate the curve at equidistant phases $t_i=\tfrac{i}{H}$, $i=1,\dots,H$.

Given supervision targets $\{x_i\}_{i=1}^H$ on the same $H$ phases, we train $\psi$ by maximizing the (weighted) log-likelihood under the corresponding Gaussian marginals. The per-step log-likelihood is
\begin{align*}
\ell_{\psi,i} = &-\frac{1}{2}\left(\mu_\psi(t_i) - x_i\right)^T \Sigma_\psi(t_i)^{-1} \left(\mu_\psi(t_i)-x_i\right)\\
& -\frac{1}{2} \log \det (\Sigma_\psi(t_i))\\
&+ C,
\end{align*}
where $C$ is constant w.r.t.\ the parameters. With $N$ training trajectories $\{x^{(j)}\}_{j=1}^N$, we use a recency-weighted objective (with $\lambda_\psi\!\in\!(0,1]$) over the horizon:
\begin{equation}
\mathcal{L}_{\psi}
= \frac{1}{N}\sum_{j=1}^N
\frac{\sum_{i=1}^H \lambda_\psi^{\,i-1}\,\ell_{\psi,i}\!\bigl(x^{(j)}_i\bigr)}
{\sum_{i=1}^H \lambda_\psi^{\,i-1}}.
\label{eq:loss_prediction}
\end{equation}
In the next section, we leverage the predicted future trajectories for reward shaping.

\subsection{Contextual Markov Decision Process for Racing}
In this section, we elaborate on our domain-informed design of components for the Contextual Markov Decision Process (CMDP) for the setting of racing. 

\begin{figure}[t]
    \centering
        \setlength{\abovecaptionskip}{1pt}  
    \setlength{\belowcaptionskip}{2pt}
    \includegraphics[trim=.7cm .3cm .2cm 0cm, clip]{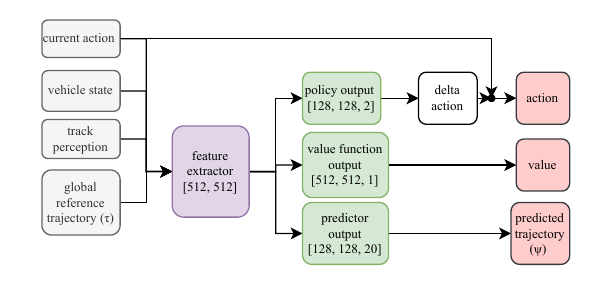}
    \vspace{-15pt} 
    \caption{Overview of the policy structure with the components and their corresponding dimensions.}
    \vspace{-10pt} 
    \label{fig:overview}
\end{figure}

\textbf{Actions $a_t$}:
As in previous work \cite{ju_digital_twin_2023, sonyAI_2022}, actions for race driving are defined as $a_t=[\alpha_t, \delta_t]$, combining acceleration and brake by $\alpha = \alpha_\text{acc} - \alpha_\text{brake}$, together with steering wheel angle $\delta_t$. However, in this work, actions are parameterized as relative changes $\Delta a_t$ from the current actions. As illustrated in Fig. \ref{fig:overview}, the policy network predicts these relative changes, with the current absolute actions included as part of the state space.
Using relative actions may help the agent focus on learning the underlying dynamics of driving rather than relying on static associations between states and absolute actions.
The relative actions are parameterized using a Gaussian distribution with state-independent but learnable diagonal covariance $\Sigma_\pi$ and state-dependent mean $\mu_\pi(s_t)$, given by a neural network. The probability density function over relative actions for a given state is then defined as ${\pi(\Delta a_t | s_t) \sim \mathcal{N}(\mu_\pi(s_t), \Sigma_\pi)}$. Then, absolute actions are obtained by ${a_t = \text{clip}(a_{t - 1} + \Delta a_t,\,a_\text{min},\,a_\text{max})}$. Clipping ensures that the action values stay within valid ranges.

\textbf{States $s_t$}:
As illustrated in Fig. \ref{fig:overview}, the inputs to our policy are divided into four essential groups: 
\begin{itemize}
    \item Vehicle states: absolute velocity ($v$), longitudinal acceleration ($a_x$), lateral acceleration ($a_y$), average front/rear slip ratio ($r_F$, $r_R$), average front/rear slip angle ($\beta_F$, $\beta_R$)\footnote{We adopt the standard definitions of slip angles and slip ratios as in \cite{milliken:1995:RCVD}.}. These vehicle states are essential for race driving to let the agent capture the current state of the car when pushing it to the performance optimum. Note that these states are also perceivable by human race drivers. 
    \item track perception: track boundary points as in \cite{ju_digital_twin_2023, sonyAI_2022} are adopted here. It contains a series of track border points on both sides of the track boundaries at distances \{5, 10, 20, 40, 80, 160, 320\} relative to the current position of the center of gravity of the car, transferred into the car local coordinates framework. 
    \item Global reference trajectory waypoint. We feed waypoints of the global reference trajectory $\tau$ as features, departing from previous work \cite{ju_digital_twin_2023, lockel_adaptive_2023} where local path planning features were employed. The waypoints are given on a 5-second horizon with 0.25 seconds in between and converted into the car's local coordinates as well. 
    \item Current actions. As we are using relative actions and the agent needs to know the current position of the pedals and steering wheel, they are fed into the policy as well. 
\end{itemize}

\textbf{Policy $\pi$}:
The network structure is shown in Fig. \ref{fig:overview} with the dimensions of each module. A two-layer feedforward net extracts features from the inputs and feeds the latent variables to the policy network, which predicts the relative actions, the value function network, and the trajectory predictor output, which will be discussed in the following section.  

\textbf{Context $c$}:
According to our goal of imitating trajectories of expert drivers, we choose spatial trajectories as the context of our MDP.
We assume that the trajectories $\tau \subset \mathbb{R}^d$ follow a distribution $\tau \sim \mu_\tau$. While $\mu_\tau$ could be the sampling distribution of expert data, we will lay out a data augmentation approach by learning a distribution in Section \ref{sec:policy_training}.

\begin{figure}[tb]
    \centering
        \setlength{\abovecaptionskip}{1pt}  
    \setlength{\belowcaptionskip}{2pt}   \includegraphics[width=.7\columnwidth, trim=0cm 0.2cm 0cm 0cm, clip]{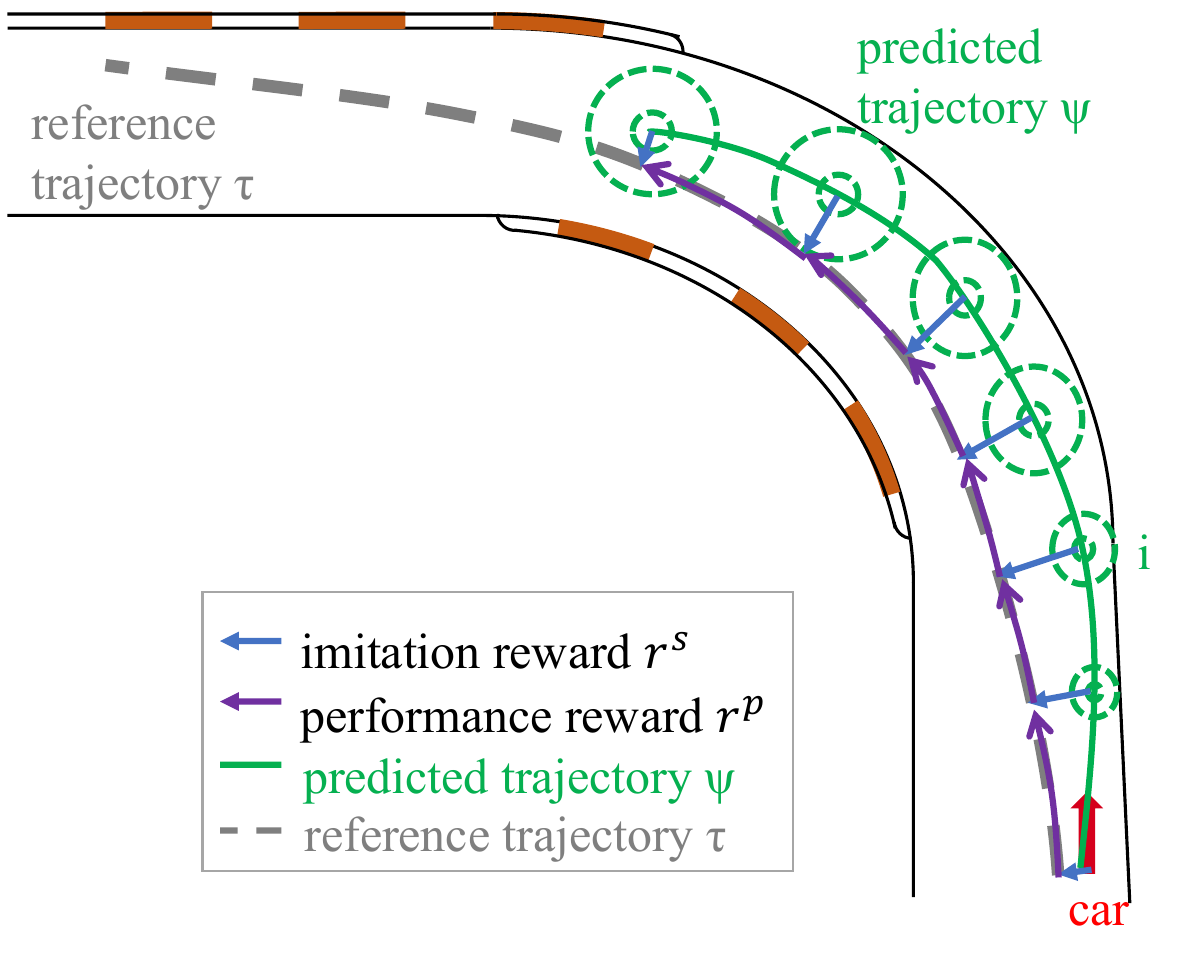}
    \vspace{-10pt} 
    \caption{A schematic diagram for reward terms. }
    \label{fig:reward}
\end{figure}

\textbf{Reward}: Here, we will define the performance reward $r^p$, style imitation reward $r^s$, and their extensions to the receding horizon prediction $r^p_\psi$ and $r^s_\psi$ respectively. We show a schematic diagram of the reward terms in Fig. \ref{fig:reward}.

Since we define the rewards based on the context reference trajectory $\tau$, we introduce a projection operator
$\phi_\tau: \mathcal{S}\to\mathbb{R}^d$.
The operator $\phi_\tau$ maps a given state or a prediction to the closest point on the reference trajectory $\tau$.
Using $\phi_\tau$ in the reward will not allow us to compute expected values in closed form. Instead, we apply Monte Carlo sampling to approximate the expected values of the look-ahead rewards.

\paragraph*{Performance Reward} In our race car setting, performance can be measured in terms of distance covered along a reference trajectory per time step.
For a given reference $\tau$ and two points $x$ and $y$ on $\tau$, we denote the absolute arc-length distance between the points as $d_\tau(x,\ y)$ and use it to compute the progress reward
\begin{equation}
    r^p(s_t) = d_\tau\left(\phi_\tau(s_{t+1}),\ \phi_\tau(s_t) \right).
\end{equation}

In addition, we add penalties for unhealthy states to the progress reward, which trigger the early termination of an episode, including off-track, car spin, and slow speed. 
Along the predicted trajectory, we can take the expected progress between succeeding points. However, due to linearity of expectation and the additivity of arc-length, this simplifies to the expected overall progress along the predicted horizon. We do not use a scaling factor for this reward. Therefore, we obtain:
\begin{align}
r^p_\psi(s_t) &= \sum_{i=1}^H\ \mathbb{E}_{\psi_i} \Bigr[ d_\tau \bigl( \phi_\tau(\psi_i),\ \phi_\tau(\psi_{i-1}\bigr) \Bigl] \nonumber \\
&=\mathbb{E}_{\psi_H}\Bigr[ d_\tau \bigl( \phi_\tau(\psi_H),\ \phi_\tau(s_t)\bigr)\Bigl].
\end{align}

\paragraph*{Style Imitation Reward}
The imitation reward is designed to encourage the agent to closely follow the reference trajectory $\tau$, guiding it to optimize performance in a manner that mirrors human behavior. Following the approach in \cite{deepMimic}, we decouple the reference trajectory from time and use projected distances. This choice mitigates issues associated with accumulated errors in time alignment or domain shifts between collected data and our simulation environment.
To determine the alignment of a state of predicted future points with the reference trajectory, we consider a style imitation distance function $d_s(x,\ y)$, which measures the distance between a sampled predicted point $x$ and its projection onto the reference $y=\phi_\tau(x)$. For our application, we choose the Euclidean distance.

Based on this distance function and the projection operator, we can define the style imitation reward by taking the negative exponential of the style imitation distance:
\begin{equation}
    r^s(s_t) = \text{exp}\left(-\alpha_d\,d_s \left( s_t,\ \phi_\tau(s_t)\right)^2\right).
\end{equation}
Extending this reward to the predicted trajectory is straightforward by using the expected imitation distance of predicted points:
\begin{equation}
    r^s_\psi(s_t) = \sum_{i=1}^H\ \alpha_\psi^i\ \mathbb{E}_{\psi_i}\biggl[\text{exp}\Bigl(-\alpha_d\ d_s\bigl(\psi_i,\ \phi_\tau(\psi_i)\bigr)^2\Bigr)\biggr].
\end{equation}
Overall, we obtain the combined reward 
\begin{equation}
    r(s_t) = r^p(s_t) + r^p_\psi(s_t) + \alpha^s\,\bigl(r^s(s_t) + r^s_\psi(s_t)\bigr).
\end{equation}

\subsection{Pretraining and Policy Fine-Tuning with Data Augmentation}
\label{sec:policy_training}
In this section, we detail the training pipeline, comprising (i) supervised pretraining of the policy and the auxiliary trajectory predictor, and (ii) reinforcement learning fine-tuning with PPO under data augmentation drawn from the reference trajectory distribution (RTD). Pretraining provides stable initialization and improves sample efficiency, while RTD-based augmentation broadens coverage of feasible lines and setups during fine-tuning. The complete procedure is summarized in Algorithm~\ref{alg:main}.

\textbf{Data Augmentation}: Instead of using trajectories of expert demonstrations directly as context, we first fit a distribution to these trajectories called \textit{Reference Trajectory Distribution (RTD)} using \textit{Probabilistic Movement Primitives (ProMP)} \cite{promp}, similar to \cite{ju_digital_twin_2023}. 
The samples drawn from this distribution are then used as context for policy training. 
This step is particularly helpful to enhance training robustness when the amount of human demonstrations is limited, and even with adequate demonstration data, data augmentation improves generalization and performance.
We consider a distance-based representation of the trajectories to capture spatial correlations efficiently.
Then, using regression with radial basis functions, we reduce the dimension of each expert trajectory. 
In this low-dimensional space, we can efficiently fit a Gaussian distribution $\mathcal{N}\left(\mu_w,\,\Sigma_w\right)$ to the weight coefficients $w_i$ of the trajectories, where $\mu_w$ and $\Sigma_w$ are determined as the sample mean and sample covariance respectively.
During rollout generation, reference trajectories are drawn from the RTD by sampling $w^\tau \sim \mathcal{N}\left(\mu_w,\,\Sigma_w\right)$ and using it as weights for the radial basis functions to recover the full trajectory $\tau$. 
By utilizing RTD-based data augmentation, an arbitrary number of human-like reference trajectories can be generated for policy training, enhancing the diversity of training data while maintaining human-like characteristics.

\textbf{Pretraining}: Before policy training using PPO, the policy and trajectory predictor are jointly pretrained using supervised learning.
As in previous works \cite{benchmark_IL_racing, ju_digital_twin_2023}, we use expert demonstrations to train the policy with Behavioral Cloning. Additionally, we use the corresponding expert trajectories to train the trajectory predictor. Both components are trained jointly using a dataset $\boldsymbol{D}_\text{pt}=\left\{(s_i, \Delta a_i,\,x^{(i)}),\, i=1,\dots,N_{\text{pt}}\right\}$ consisting of states and corresponding relative actions and expert trajectories $x^{(i)}$, where $N_\text{pt}$ refers to the total number of data points for pretraining.
Due to our parametrization of actions, Behavioral Cloning reduces to minimization of the mean-squared error between expert relative actions and mean values predicted by the policy.
Experimentally, we found that adding a regularization term with the coefficient $\alpha_{\text{reg}}$ to the mean squared error BC loss improves learning in the early stages of consecutive application of PPO.
Together with the loss $\mathcal{L}_{\psi}$ of the trajectory predictor with a scaling coefficient $\alpha_{\psi, \text{pt}}$, we obtain the combined objective for pretraining $\mathcal{L}_{\text{pt}}(\boldsymbol{D}_\text{pt})$ as the average on the dataset $\boldsymbol{D}_\text{pt}$:
\begin{align}
    \mathcal{L}_{\text{BC}}(\boldsymbol{D}_\text{pt}) &= \frac{1}{N_{\text{pt}}} \sum_{i=1}^{N_{\text{pt}}}(\Delta a_i - \mu_\pi(s_i))^2\ + \alpha_{\text{reg}}\,\mu_\pi(s_i)^2\\
    \mathcal{L}_{\text{pt}}(\boldsymbol{D}_\text{pt}) &= \mathcal{L}_{\text{BC}}(\boldsymbol{D}_\text{pt}) - \alpha_{\psi, \text{pt}}\ \mathcal{L}_{\psi}(\boldsymbol{D}_\text{pt}).
    \label{eq:loss_pretrain}
\end{align}

\textbf{Policy Fine Tuning}: We train the policy with PPO, which uses as alternating scheme of collecting data by performing rollouts with the current policy and successively updating the policy based on the collected data.
Before updating the policy, we propose first to update the trajectory predictor, and then calculate rewards based on predicted trajectory $\psi$. 
Hence, we always consider the most recent policy for the receding horizon predictions.
We also found it crucial to use early stopping while training the predictor to prevent overfitting to individual trajectories. 

For policy optimization, we follow the standard PPO procedure, using a clipped surrogate objective and entropy regularization. 
All components are optimized jointly using gradient ascent on a combined optimization objective. Here, training of the predictor is included as an auxiliary task to provide the known benefits of improving the learned representation, resulting in the following maximization objective for policy training
\begin{equation}
    \mathcal{L}_\pi = \mathbb{E}_{\pi} \left[\mathcal{L}_{\text{PG}} + \mathcal{L}_\text{H} - \mathcal{L}_{\text{VF}} + \mathcal{L}_{\psi} \right]
    \label{eq:loss_pi}
\end{equation}
where $\mathcal{L}_{\text{PG}}$, $\mathcal{L}_\text{H}$, $\mathcal{L}_{\text{VF}}$ represent policy gradient loss, entropy regularization loss and value function loss of the learned value function $V$, respectively. $\mathcal{L}_{\psi}$ is defined in Equation \ref{eq:loss_prediction}. For more details regarding the PPO loss items, we refer to \cite{PPO}. After the policy update, we update the reward coefficient $\alpha^s$ using Equation \ref{eq:alpha}. The resulting overall training scheme is laid out in Algorithm \ref{alg:main}.



\begin{figure}
    \begin{minipage}{\columnwidth}
    \begin{algorithm}[H]
    \footnotesize
    \caption{Proposed Training Algorithm}
    \label{alg:main}
    \begin{algorithmic}
    \State Generate RTD from expert demonstrations
    \State Initialize $\pi$ and $\psi$ with supervised prelearning loss $\mathcal{L}_{\textbf{pt}}$ in Eq. \ref{eq:loss_pretrain}
    
    \For {each policy update} 
    \State Collect rollout data
    \State Update $\psi$ with $\mathcal{L}_{\psi}$ in Eq. \ref{eq:loss_prediction}
    \State Compute $r^p_\psi$ and $r^s_\psi$ with predictions from updated $\psi$
    \State PPO policy update with $\mathcal{L}_\pi$ in Eq. \ref{eq:loss_pi}
    \State Update coefficient $\alpha^s$ with $\mathcal{L}_{\alpha^s}$ in Eq. \ref{eq:alpha}
    \EndFor 
    \end{algorithmic}
    \end{algorithm}
    \end{minipage}
\end{figure}

\section{EXPERIMENTS AND EVALUATION}
\label{sec:experiments}
We conducted a series of experiments to evaluate the performance of our method, focusing on its core components: the receding horizon predictive reward, the prediction module, and data augmentation.

The training was performed using an in-house high-fidelity race car simulation model.
All experiments were conducted on either a desktop computer or a CPU cluster, utilizing 8 parallel simulation environments. For pretraining and data augmentation, we use demonstrations from a dataset collected by professional race car drivers, encompassing various car configurations and tracks. Due to computational resource constraints, our evaluation initially focused on a single track to thoroughly examine the method and perform ablation studies. To demonstrate the effectiveness and generalization capability of our method, we extend the primary evaluation to additional car configurations and tracks, comparing its performance against baseline methods.

The following sections provide a detailed analysis of the experiments, including performance comparisons with baseline methods and ablation studies to quantify the contribution of individual components of our method.

\subsection{Method evaluation and benchmarking}
\begin{table*}[tbh]
\centering
\caption{Experiment definition and results.}
\label{tab:exp_def}
\begin{tabular}{l|l|l|l|l|l|l}
\hline
\textbf{exp} & \textbf{prediction} & $\boldsymbol{r^s}$ & $\boldsymbol{r^p}$ & $\boldsymbol{\alpha^s}$ & \textbf{lap time (Track A)} & \textbf{MRO (Track A)} \\ \hline
A ours  & yes & $r^{s,t}, r^{s, \psi}$ & $r^{p,t}, r^{p, \psi}$ & $\hat R=0.009$ & \textbf{66.229} $\pm$ \textbf{0.0259} & \textbf{0.2891} $\pm$ \textbf{0.0052}\\ \hline
B step  & no & $r^{s,t}$ & $r^{p,t}$ & 1 & $66.961 \pm 0.299$ & $0.4547 \pm 0.0320$\\ \hline
C adv  & no & $r^{s, \text{ad}}$ & $r^{p,t}$& 0.1 & \textbf{66.775} $\pm$ 0.0143 & $\textbf{0.1341} \pm \textbf{0.0035}$  \\ \hline
D CRL  & no & 0 & $r^{p,t}$ & 0 & 67.209 $\pm$ 0.275 & $0.7686 \pm 0.0098$\\ \hline
\end{tabular}

\end{table*}


In this section, we evaluate the proposed method's general performance compared with baselines. On track A, each setting is tested on one environment with 20 seeds. We further elaborate on the effect of multiobjective learning using the different methods despite their different performance-imitation trade-offs. Table \ref{tab:exp_def} presents the experimental setup and baseline comparisons, evaluated using two key performance metrics: lap time and mean reference offset (MRO)\footnote{Mean Reference Offset (MRO) is defined as the average of the absolute value of the lateral distance to the reference trajectory.}. These metrics measure both the performance of the method and its ability to imitate human driving behavior. Given our objective of optimizing performance, we aggregate the minimum lap time during training over all seeds and their corresponding mean reference offset, with the mean and standard error as the confidence interval.

It is important to highlight that all experimental settings are trained with data augmentation based on the Reference Trajectory Distribution (RTD) to avoid its influence on the learning process. While our approach incorporates data augmentation, it is not originally included in the methods used for Experiment Settings B and C \cite{deepMimic, amp}. A dedicated evaluation of the impact of data augmentation is presented in Section \ref{sec:eva_ablation}. We evaluate the following methods, each using 100 million steps with a total of 1,525 policy updates:

\begin{itemize}
\item A-ours: stepwise and predicted performance and imitation reward, trajectory prediction as an auxiliary task, trained with data augmentation. 
\item B-step reward: Stepwise performance and imitation reward. This is adopted from previous work (e.g. \cite{deepMimic}, which is widely used for trajectory following problems for robots). To make it a clear comparison for the reward design, both rewards are calculated on the phase-based reference trajectories and trained with augmented data, which is shown helpful via further ablation studies in Section \ref{sec:eva_ablation}\footnote{We did a grid search for $\alpha^s=0.2, 0.4, 0.6, 0.8, 1$, and with $\alpha^s=1$ it achieves best lap time and imitation. Full results are listed in \ref{app:amp}}. 
\item C-adversarial reward (AMP): Stepwise performance and adversarial imitation reward, adopted from Adversarial Motion Priors (AMP)~\cite{amp}, which is a well-known imitation learning designed to imitate style while optimizing task objectives. Same with for Baseline B, we also train it with data augmentation here\footnote{It is challenging to tune the coefficient for the discriminator reward and we found that the imitation-performance trade-off can not be effectively controlled by the coefficient. In Table \ref{tab:exp_def} we show results with $\alpha^s=0.1$. More results are listed in Appendix \ref{app:amp}.}.  
\item D-CRL: Contextual reinforcement learning for racing\cite{ju_digital_twin_2023} with stepwise performance reward calculated on the reference trajectory. Unlike B and C, which are used mainly for robots, this method is tailored on a similar experimental setting for high-fidelity race car simulation.
\end{itemize}

\subsubsection{Learning Progress}
\begin{figure}[tbh]
    \centering
    \subfigure[Lap Time]{
        \includegraphics[trim=0cm 0.25cm 0cm 0cm, clip]{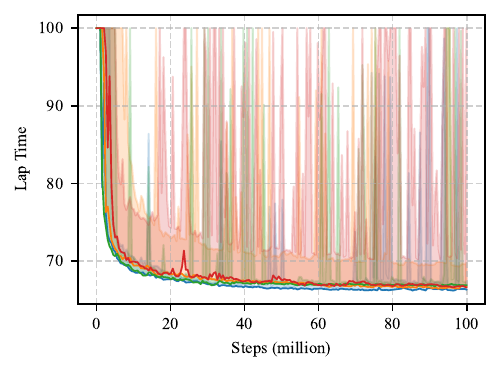} 
        \label{fig:learning_laptime}
    }
    \subfigure[Mean Reference Offset]{
        \includegraphics[trim=0cm 0.25cm 0cm 0.2cm, clip]{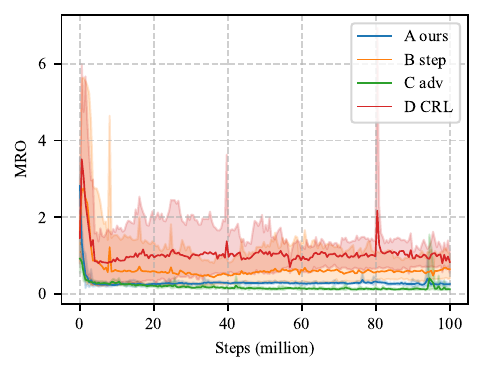} 
        \label{fig:learning_MRO}
    }
    \vspace{-8pt}
    \caption{Learning curves of Lap Time and Mean Offset of Experiment B, C, and D, compared with ours.  For lap time, we plot the minimum value over the seeds, as the objective is to optimize performance. For MRO, we plot the average value across the seeds to assess imitation quality. The shaded area in both plots represents the range covered the seeds, providing insight into the variability of training outcomes. }
    \label{fig:exp:DRH_ABCD}
\end{figure}

The learning progress of the four experiments, each trained with five different seeds, is shown in Fig. \ref{fig:exp:DRH_ABCD}. Throughout the early training stages, all experimental settings struggle to complete a full lap. These failures, represented by spikes reaching up to 100, signify cases where the agent made a critical mistake, such as losing control and spinning out or going off-track, leading to an early termination of the rollout. However, as training progresses, particularly after approximately 5 million steps, the agents begin to complete full laps more consistently, gradually improving their driving behavior. Despite this improvement, distinct differences emerge between the various experimental setups. Both Experiment B and Experiment D demonstrate difficulty in maintaining imitation quality as lap time optimization becomes the primary focus. The imitation performance deteriorates, suggesting that the agents prioritize speed at the expense of mimicking human-like driving behavior. Conversely, Experiment C focuses on imitation fidelity, successfully improving the MRO, yet it struggles to further optimize lap time beyond approximately 50 million steps. This indicates that while it effectively follows the reference trajectories, it lacks the necessary adaptation to improve overall driving performance. In contrast, our method effectively balances imitation and performance. By incorporating imitation constraints, it ensures that the agent maintains a predefined level of imitation quality while continuously optimizing lap time. This approach enables the agent to explore performance improvements without deviating excessively from the reference behavior, ultimately achieving a well-rounded trade-off between performance and human-like driving characteristics.

\subsubsection{Multiobjective Optimization}
\begin{figure}
    \centering
    \includegraphics{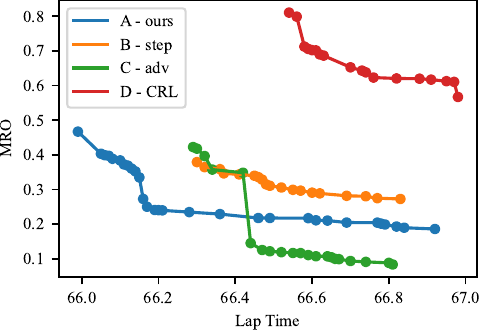}
    \vspace{-10pt}
    \caption{The Pareto fronts for different methods. Since both lap time and Mean Reference Offset (MRO) need to be minimized, the optimal solutions lie in the lower-left corner of the plot. Each point represents a lap from the training epochs that lie on the Pareto front within its respective experiment group. For improved readability, we have omitted data points that are not on the Pareto front.}
    \label{fig:exp_pareto}
\end{figure}
Experiments B and C require balancing the trade-off between performance and imitation.
Unlike conventional approaches that necessitate manual tuning of the trade-off coefficient, our method achieves this by setting an imitation target based on predefined imitation requirements.

To further assess the capability of the different methods to learn optimal trade-offs between the conflicting objective of style-biased reinforcement learning, we evaluate each method for a grid of different trade-off coefficients (or imitation bounds) and visualize their Pareto frontier, by showing all non-dominated policies that were found for each respective method. The Pareto froniters are shown in Fig.~\ref{fig:exp_pareto}, where each point represents a lap within its respective experiment group, selected based on its dominant lap time and Mean Reference Offset (MRO). Given that the lap time difference between top professional drivers is exceptionally small (typically less than 0.2 seconds between the first and second positions in qualifying), we omit data points where the lap time exceeds the fastest recorded lap by more than one second. The imitation coefficient values selected for the results are as follows: Setting A: $\hat R =$ 0.85, 0.87, 0.89, 0.91, 0.93, 0.95, 0.97. Setting B: $\alpha^s=$ 0.2, 0.4, 0.6, 0.8, 1. Setting C: $\alpha^s=$ 0.01, 0.03, 0.05, 0.07, 0.09, 0.1, 0.3, 0.5, 0.7, 0.9, 1\footnote{We found it difficult to tune the coefficient and conducted a second grid search from 0 to 0.1 after an initial search from 0 to 1. More results are provided in Appendix \ref{app:amp}.}.

From the results, we observe that Setting D with CRL exhibits the worst performance in imitation, which is expected since it does not explicitly encourage trajectory imitation, apart from calculating the performance reward based on the reference trajectories. Setting B demonstrates improved performance and imitation, but does not dominate either metric when compared to Setting A and C. Setting C, which uses an adversarial reward, achieves the best imitation performance, but tuning the imitation-performance trade-off is particularly challenging. Despite performing two rounds of grid search, we were unable to achieve the top performance observed with our method. This could be attributed to discrepancies between the demonstration data and the testing environment, making it difficult for the adversarial reward to match the state distribution while simultaneously optimizing performance. In the top-performance area (66.0 to 66.4), our method dominates all other baselines. 

\subsubsection{Generalization}
\begin{table*}[tb]
\centering
\caption{Experiments for track B and C}
\label{tab:exp_2track}
\begin{tabular}{l|l|l|l|l}
\hline
\textbf{exp} & \textbf{lap time (B)} & \textbf{MRO (B)} & \textbf{lap time (C)} & \textbf{MRO (C)} \\ \hline
A ours & \textbf{66.892} $\pm$ 0.0256 & \textbf{0.2417} $\pm$ 0.0166 & \textbf{69.688} $\pm$ 0.0473 & \textbf{0.2989} $\pm$ 0.0109 \\ \hline
B step &  \textbf{67.060} $\pm$ 0.0318 & 0.2743 $\pm$ 0.0208 & \textbf{69.318} $\pm$ 0.0510 & 0.7296 $\pm$ 0.0976 \\ \hline
C adv &  67.402  $\pm$ 0.0215 & \textbf{0.0789} $\pm$ 0.0054 & 69.822 $\pm$ 0.0365 & \textbf{0.1535} $\pm$ 0.0136 \\ \hline
D CRL & 67.458 $\pm$ 0.2242    & 0.7053 $\pm$ 0.0096 & 69.910 $\pm$ 0.1182 & 0.9703 $\pm$ 0.0101\\ \hline
\end{tabular}
\end{table*}

To demonstrate the generalizability and applicability of our approach, we further evaluate our method against baseline approaches on two additional race tracks. All three tracks are official world championship circuits, known for their complexity and the challenge they pose in achieving optimal performance.
For Experiment B and C, we use the same $\alpha^s$ values as in Table \ref{tab:exp_def}, which were optimized for Track A. Due to computational constraints, each experiment was run with five different seeds, and the results are summarized in Table \ref{tab:exp_2track}.
Importantly, we use the same imitation-performance trade-off coefficient for all tracks. The results show that for Experiment Settings B and C, the achieved performance and imitation quality vary across tracks, suggesting that these methods require individual tuning for each environment. In contrast, our method consistently delivers top performance while maintaining imitation requirements, demonstrating its robustness across different tracks without the need for manual tuning.

\subsection{Performance analysis}
\subsubsection{Single lap compared to references.} 
\begin{figure}
    \centering
    \includegraphics[width=.85\columnwidth]{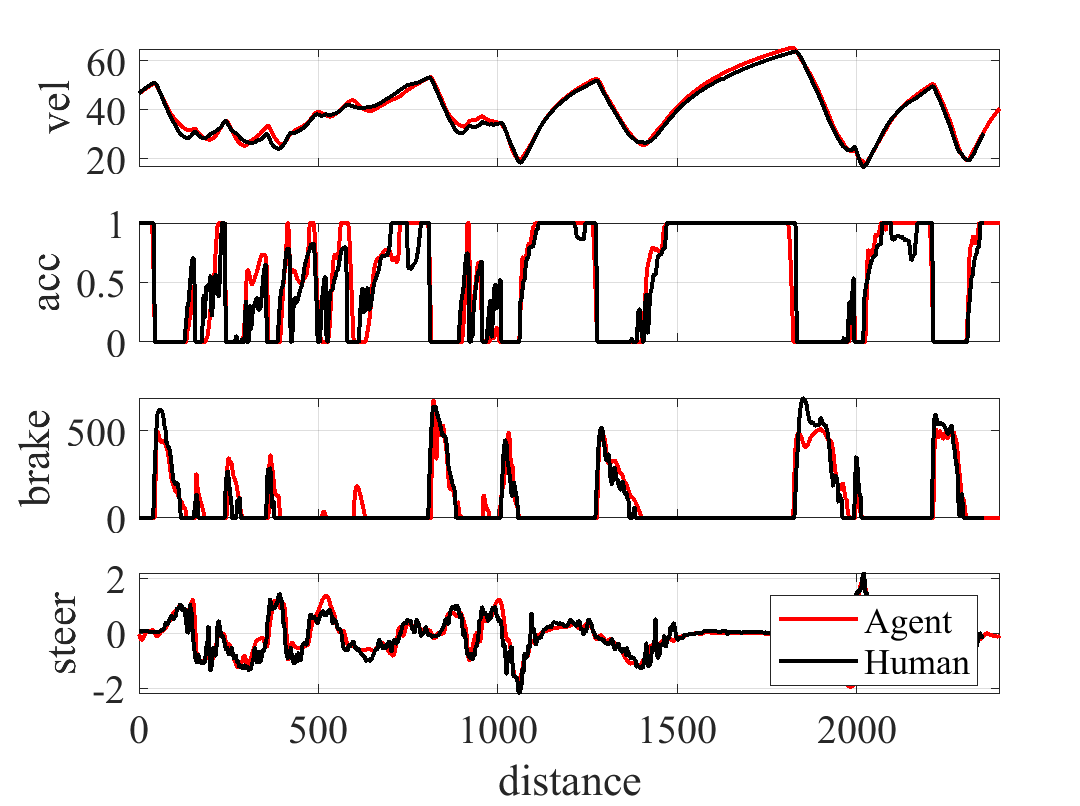}
        \vspace{-10pt}
    \caption{Single lap compared to the reference trajectory.}
    \label{fig:exp_single_lap}
\end{figure}

We evaluate a single agent-driven lap by comparing its time-series data with the reference lap from human demonstrations, as shown in Fig. \ref{fig:exp_single_lap}. The figure presents the speed profile along with the three control actions (throttle, brake, and steering) for both the agent and the reference lap. With an MRO of 0.34m, the agent follows the reference trajectory closely, exhibiting similar speed and action profiles. However, it optimizes performance by correcting suboptimal human actions caused by disturbances on the track. For example, at approximately 750m and 2100m, the human driver lifts off the throttle before braking. In contrast, the agent gains time by maintaining throttle input and executes a more optimal braking strategy, demonstrating its ability to refine human demonstrations for improved performance.

\subsubsection{Trajectory imitation}
\begin{figure}
    \centering
    \includegraphics[width=\columnwidth, trim=0cm 0.25cm 0cm 0.2cm, clip]{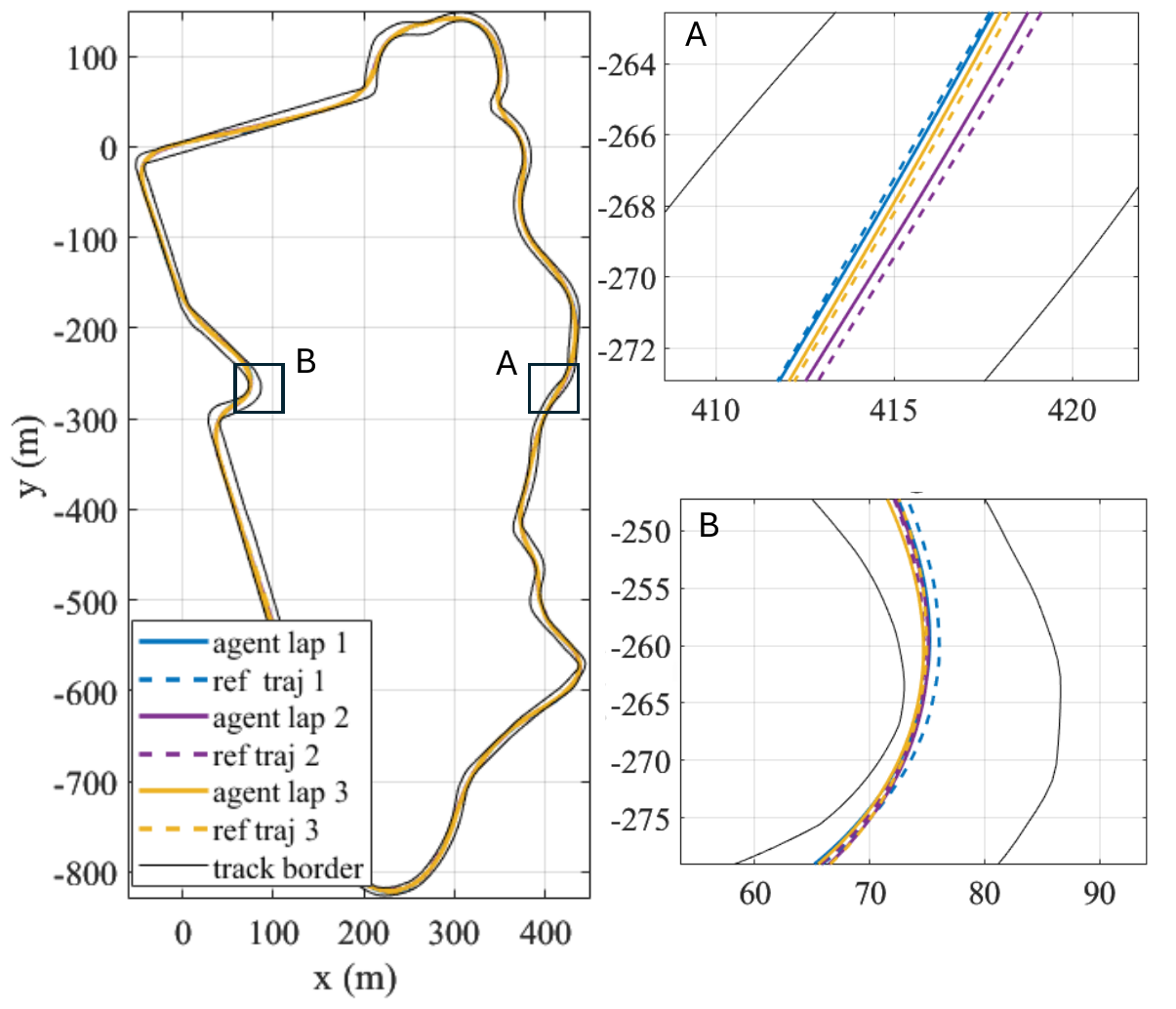}
    \vspace{-20pt}
    \caption{Driving lines of agent laps and their corresponding reference trajectories.}
    \label{fig:exp_line_imitation}
\end{figure}

Fig. \ref{fig:exp_line_imitation} illustrates how the agent follows the reference trajectories at different locations on the track using three randomly sampled reference trajectories. In most cases, the agent closely follows the given reference trajectory (e.g. A). However, in certain areas, particularly mid-corner (e.g., location B), the agent may deviate slightly. This deviation is a result of performance optimization learned through the performance reward. Since the reference trajectories are randomly sampled, they may not always be optimal or entirely consistent. Consequently, the agent learns to make minor local deviations from the reference line to achieve better overall performance.

\subsection{Ablation Study}
\label{sec:eva_ablation}
\subsubsection{Automatic reward scaling factor tuning (S)}
\begin{figure}[ht]
    \centering
    \setlength{\abovecaptionskip}{2pt}  
    \setlength{\belowcaptionskip}{10pt}
    \subfigure[imitation reward (training)]{
    \vspace{-5pt}
        \includegraphics[width=0.48\columnwidth]{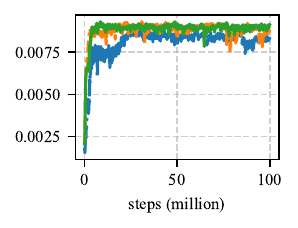} 
    \label{fig:sub1}
    }
    \subfigure[style coefficient $\alpha^s$ (training)]{
            \includegraphics[width=0.42\columnwidth]{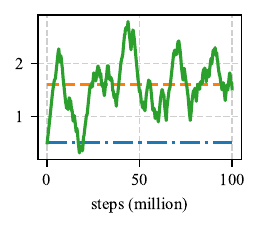} 
        \label{fig:sub2}
    }
    \subfigure[lap time (evaluation)]{
        \includegraphics[width=0.47\columnwidth]{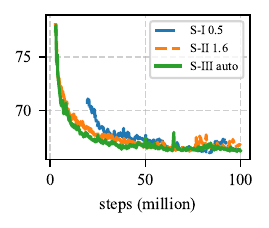} 
        \label{fig:sub1}
    }
    \subfigure[mean ref offset (evaluation)]{
    \vspace{-15pt}
        \includegraphics[width=0.46\columnwidth, trim=0cm 0cm 0.2cm 0.1cm, clip]{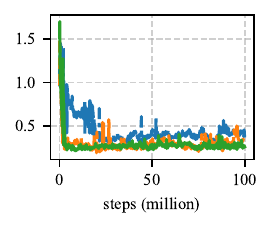} 
        \label{fig:sub2}
    }
    \vspace{-8pt}
    \caption{Learning progress of experiment S-I, S-II und S-II, using reward coefficient $\alpha^s$ 0.5, 1.6 and automatic coefficient tuning with imitation target $\hat R = 0.9$}
    \label{fig:exp_scaling}
\end{figure}

In this section, we present the result and benefit of our automatic reward coefficient scaling for the performance-imitation trade-off. Our experiment settings are as follows: 
\begin{itemize}
    \item S-I: fixed scaling with $\alpha^s = 0.5$
    \item S-II: fixed scaling with $\alpha^s = 1.6$
    \item S-III: auto scaling with $\hat R = 0.009$ which approximately corresponds to MRO $= 0.325$ m . 
\end{itemize}
The results are summarized in Table \ref{tab:exp_scaling}, while the learning progress, including the batch-averaged imitation reward, the style reward coefficient $\alpha^s$, lap time, and Mean Reference Offset (MRO) during evaluation runs, are depicted in Fig. \ref{fig:exp_scaling}. We first conducted S-III with automatic reward scaling and then selected fixed coefficients based on the result for S-I and S-II to evaluate their effects. Increasing the scaling coefficient from 0.5 to 1.6 (S-I vs. S-II) shifts the trade-off towards style imitation, leading to a performance decline while improving imitation quality by approximately one-third. By setting an imitation target and employing automatic coefficient tuning (S-III), the reward coefficient is dynamically adjusted during training, achieving a similar imitation quality to S-II while maintaining performance close to S-I.

\begin{table}[tb]
\centering
\caption{Experiments for automatic scaling}
\label{tab:exp_scaling}
\begin{tabularx}{\columnwidth}{X|l|X|X}
\hline
\textbf{exp} & \textbf{scaling} & \textbf{lap time} & \textbf{mean ref offset}  \\ \hline
S-I: fixed scaling & 0.5 & $\textbf{66.120} \pm 0.0216$ & $0.3799 \pm 0.0243$  \\ \hline
S-II: fixed scaling & 1.6 & $66.226 \pm 0.0631$& $\textbf{0.2732} \pm 0.0070$ \\ \hline
S-III: auto scaling & auto & $\textbf{66.163} \pm 0.0229$ & $\textbf{0.2629} \pm 0.0096$  \\ \hline
\end{tabularx}
\end{table}

These findings highlight the advantages of our automatic reward tuning mechanism:
a) It enables the achievement of a predefined imitation target without requiring extensive parameter searches for the scaling coefficient.
b) By dynamically guiding exploration based on the current learning status, it facilitates improved performance compared to using a fixed scaling factor, which may not be optimal throughout training.

\subsubsection{Data augmentation (DA)}
\begin{table}[tb]
\centering
\caption{Ablation experiment for data augmentation}
\label{tab:exp_data_augmentation}
\begin{tabularx}{\columnwidth}{l|X|l|l|l}
\hline
\textbf{exp} & \textbf{DA} & \textbf{data} & \textbf{lap time} & \textbf{MRO}  \\ \hline
DA-demos & no & train  & $66.156 \pm 0.035$ & $0.2861 \pm 0.0165$  \\ \hline
DA-RTD & yes & train  & $66.078 \pm 0.040$ & $0.2934 \pm 0.0093$  \\ \hline\hline
DA-demos & no & test & $66.866 \pm 0.0257$ & $0.4279 \pm 0.0176$ \\ \hline
DA-RTD & yes & test & $\boldsymbol{66.639} \pm 0.0197 $ & $0.4587 \pm 0.0067$ \\ \hline 
\end{tabularx}
\end{table}
We evaluate the data augmentation (DA) technique using the RTD fitting from human demonstrations. To assess the generalizability of an agent trained with RTD, we split the entire demonstration dataset into a training dataset (42 laps from 7 runs) and a test dataset (14 laps from 2 runs).
For DA-demos, we train the agent using the human demonstrations directly as reference lines.
For DA-RTD, we first fit the RTD and then use samples drawn from it as references for training.
Subsequently, we evaluate the agent on both the training and test datasets. Given that the demonstration dataset is relatively large and contains diverse driving data, the training performances of DA-demos and DA-RTD are similar. However, in the test set evaluation, DA-RTD outperforms DA-demos by approximately 0.2 seconds, highlighting the improved generalizability achieved through our data augmentation technique.

\subsubsection{Prediction module as auxiliary task (P)}
We also find that by introducing the trajectory prediction module as an auxiliary task, without using it for credit assignment, learning is enhanced and performance improves. As shown in Table \ref{tab:exp_prediction}, the ablation experiments are divided into two groups: settings P-I and P-II, which both use the step performance reward $r^{p}$, but differ by whether the prediction module is used. Settings P-III and P-IV, on the other hand, incorporate both the step performance and style imitation rewards. P-II outperforms P-I in terms of lap time by approximately 0.25 seconds, with slightly better imitation, supporting our hypothesis that understanding the environment and the consequences of actions helps the agent to optimize performance. A lap time improvement of 0.2 seconds and an imitation improvement of about 30\% in terms of Mean Reference Offset (MRO) can be observed between P-III and P-IV. Notably, P-IV also outperforms P-II in lap time, indicating that incorporating demonstration data and biasing the style helps improve performance by guiding exploration using expert knowledge.
\begin{table}[tb]
\centering
\caption{Ablation experiments for prediction}
\label{tab:exp_prediction}
\begin{tabularx}{\columnwidth}{l|l|l|l|l}
\hline
\textbf{exp} & \textbf{pred}  & \textbf{reward}& \textbf{lap time} & \textbf{MRO}  \\ \hline
P I & no  & $r^{p}$ & $66.708 \pm 0.0560$ & $0.8820 \pm 0.0586$  \\ \hline
P II & yes & $r^{p}$ & $\textbf{66.454} \pm \textbf{0.0379}$ & $0.8000 \pm 0.0608$  \\ \hline\hline
P III & no & $r^{p} + r^{s}$ & $66.530 \pm 0.0560$ & $0.4984 \pm 0.0690$  \\ \hline
P IV & yes & $r^{p} + r^{s}$ & $\textbf{66.316} \pm \textbf{0.0535} $ & $\textbf{0.3512} \pm 0.0120$  \\ \hline

\end{tabularx}

\end{table}
\section{Human-Grounded Evaluation}
\label{sec:human_eval}

We evaluate the proposed framework in a high-fidelity autonomous racing environment with data collected from professional drivers. The goal of this section is to assess whether the learned policies not only achieve high task performance, but also reproduce human-consistent behavior under varying system conditions. In particular, we examine whether the model captures driver-specific responses to changes in vehicle setup, and whether it can be used as a proxy for human evaluation in a controlled experimental setting.

\subsection{Experimental protocol}

Driver-in-the-Loop (DiL) simulators are widely used in professional motorsports to evaluate the coupled driver-vehicle system under realistic conditions. In this setting, expert drivers interact with a high-fidelity simulator mounted on a Stewart platform, allowing engineers to iteratively refine vehicle setups based on both telemetry and driver feedback.

We leverage this setting as a source of expert demonstrations and as a benchmark for human-grounded validation. During a simulator session, a professional driver evaluated multiple vehicle setups on a fixed track. The driver provided qualitative feedback after each run (consisting of multiple laps), describing handling characteristics such as understeer, oversteer, and overall balance. Based on this feedback, engineers iteratively adjusted the vehicle setup.

To evaluate the proposed method, we select three setups from such a session that reflect distinct handling characteristics. For each setup, we train a separate policy using the proposed framework. Each policy is then evaluated across all three setups, generating 50 laps per setup. This results in a total of 150 laps per setup. To reduce the influence of outliers (e.g., recovery maneuvers or exploration artifacts), we follow standard practice in human-in-the-loop evaluation and retain the best 100 laps for analysis.

This protocol allows us to assess both performance and behavioral consistency across varying dynamics, and to directly compare model behavior with expert driver feedback.

\subsection{Performance and consistency analysis}

\begin{figure}
    \centering
    \includegraphics[width=.85\columnwidth]{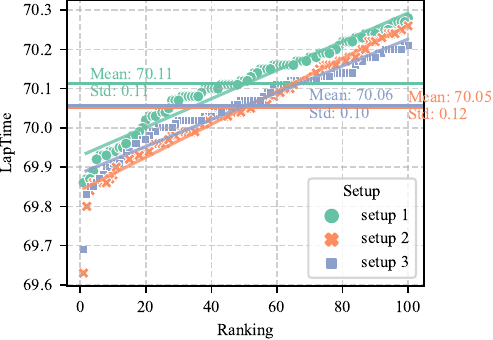}
    \vspace{-8pt}
    \caption{Lap time distributions across three vehicle setups. Each curve shows ranked lap times with linear trends indicating consistency.}
    \label{fig:usecase_ranked_laptime}
\end{figure}

We first evaluate performance using lap time distributions, shown in Fig.~\ref{fig:usecase_ranked_laptime}. Beyond best lap time, we are interested in the consistency with which high performance can be achieved. This notion of consistency reflects robustness to disturbances and sensitivity to control, and serves as a proxy for drivability in dynamic driving tasks.

The results highlight a trade-off between peak performance and consistency. While Setup 2 achieves the fastest individual lap times, it exhibits significantly higher variance, as reflected by the steeper slope of the ranked lap time curve. This indicates reduced reproducibility and greater sensitivity to small perturbations. In contrast, Setup 3 achieves slightly slower peak performance but maintains a tighter distribution, indicating more consistent behavior.

From this perspective, Setup 3 provides the best balance between performance and consistency. Notably, this ranking aligns with the qualitative assessment provided by the professional driver, who preferred setups that offered stable and predictable handling over those with higher but less controllable peak performance.

\subsection{Behavioral consistency with human drivers}

\begin{figure}
    \centering
    \setlength{\abovecaptionskip}{2pt}
    \setlength{\belowcaptionskip}{2pt}
    \includegraphics[trim=0cm 0cm 0cm 0cm, clip, width=.8\columnwidth]{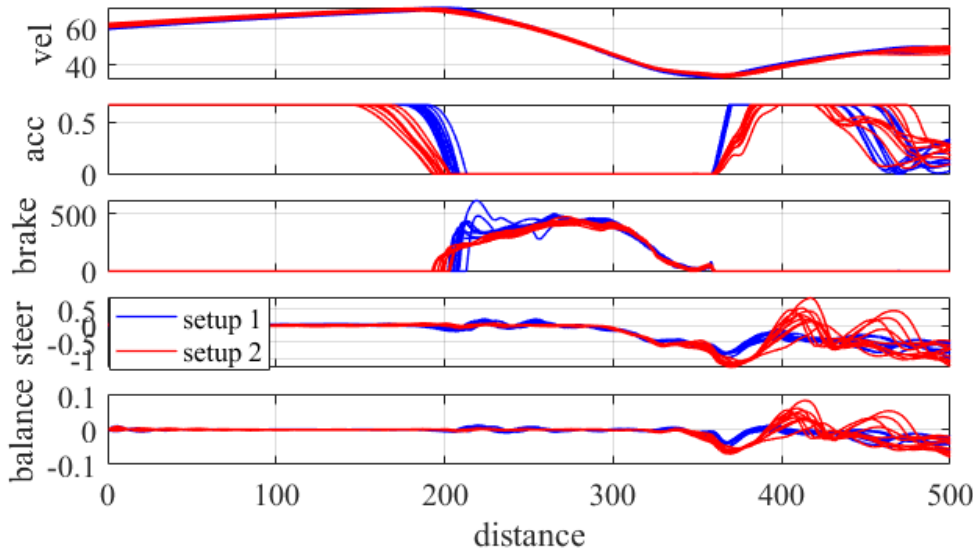}
    \vspace{-7pt}
    \caption{Time-series comparison of control and state variables for 10 laps from Setup 1 and Setup 2.}
    \label{fig:usecase_detail}
\end{figure}

To further analyze whether the learned policies capture human-like responses to setup changes, we examine time-series data of vehicle states and control inputs. Fig.~\ref{fig:usecase_detail} shows velocity, accelerator input, brake input, steering angle, and vehicle balance (defined as the difference between front and rear slip angles) for a subset of laps from two setups.

The policies exhibit systematic differences in behavior that are consistent with the underlying vehicle dynamics. For the more oversteer-prone setup (Setup 2), the policy applies earlier and smoother braking and more conservative acceleration, particularly in regions where vehicle stability is challenged. These adjustments result in higher cornering speeds but require more precise control.

In contrast, the policy trained on Setup 1 exhibits more aggressive inputs, consistent with a vehicle that requires additional rotation. These behavioral differences closely match the qualitative feedback from the professional driver, who reported insufficient rotation in Setup 1 and improved but more demanding handling in Setup 2.

This agreement suggests that the proposed framework captures not only task-level performance, but also the structure of human control strategies under varying system dynamics.

\subsection{Implications for human-referenced evaluation}

Overall, the results demonstrate that the learned policies can serve as a proxy for expert drivers in evaluating vehicle setups. The model reproduces both quantitative performance trends and qualitative behavioral differences observed in human driving. This enables controlled, repeatable experiments that retain key characteristics of human behavior, providing a practical tool for studying high-performance control systems under realistic conditions.

\section{Conclusion}
In this work, we formulate a constrained optimization problem for style-biased contextual reinforcement learning to tackle the challenge of performance optimization while maintaining imitation requirements in high-fidelity race car driving simulation. By introducing constraints on the imitation lower bound, the style imitation coefficient is dynamically tuned during learning, eliminating the need for extensive hyperparameter tuning across different environments. Additionally, the integration of receding horizon short-term prediction and lookahead reward enhances credit assignment, accelerates training, and improves final performance.

We apply this framework to race car driving, incorporating domain-informed choices of states, actions, contexts, and rewards. By leveraging pretraining and data augmentation techniques based on the reference trajectory distribution, we achieve both competitive performance and high imitation quality while generalizing effectively across different race tracks and vehicle configurations. Furthermore, we adopt this method to train a driver model for a digital twin in a driver-in-the-loop simulation, demonstrating its application in virtual setup testing for advanced motorsport development. Through a systematic and statistical analysis workflow, we validate our method using feedback from top-tier professional race drivers.

While the initial motivation stems from the demands of high-performance racing, the general formulation of optimizing task objectives while satisfying imitation constraints extends to broader robotics domains. The SBCRL framework, combined with short-term prediction and receding horizon rewards, offers a structured approach that can benefit a wide range of continuous control tasks beyond racing, paving the way for future research and applications in robotics and autonomous systems.

\section*{Acknowledgement}


The authors gratefully acknowledge the computing time provided to them on the high-performance computer Lichtenberg at the NHR Center NHR4CES@TUDa. This is funded by the German Federal Ministry of Education and Research (BMBF) and the Hessian Ministry of Science and Research, Art and Culture (HMWK).

\appendix

\subsection{Hyperparameter}
In this section, we list all relevant hyperparameters for the components, including PPO parameters in Table \ref{tab:ppo_parameters}, predictor parameters in Table \ref{tab:predictor_parameters}, rewards and automatic coefficient scaling parameters in Table \ref{tab:reward_parameters}, and pretrain hyperparameters in Table \ref{tab:pt_parameters}. 
\begin{table}[h]
    \centering
    \caption{PPO Hyperparameters}
    \label{tab:ppo_parameters}
    \begin{tabular}{l|l}
        \hline
        n\_step & 8192 \\ \hline
        num\_env & 8 \\ \hline
        $\gamma$ & 0.998 \\ \hline
        $\lambda$ & 0.98 \\ \hline
        entcoe & 1e-5 \\ \hline
        learning rate & 1e-4 \\ \hline
        activation & ReLU \\ \hline
        n\_optepoch & 4 \\ \hline
        batchsize & 1024 \\ \hline
        kl\_threshold & 0.1 \\ \hline
        vf\_coef & 1.25 \\ \hline
        H & 100 \\ \hline
        policy layers & [128, 128] \\ \hline
        feature extractor layers & [512, 512] \\ \hline
        value function layers & [512, 512] \\ \hline
    \end{tabular}
\end{table}
\begin{table}[h]
    \centering
    \caption{Trajectory Predictor Parameters}
    \label{tab:predictor_parameters}
\begin{tabular}{l|l}
        \hline
        $\lambda_{\psi}$ & 1 \\ \hline
        B\'ezier degree & 5 \\ \hline
        n\_control\_point & 10 \\ \hline
        learning rate & 1e-3 \\ \hline
        predictor layers & [128, 128] \\ \hline
    \end{tabular}
\end{table}
\begin{table}[h]
    \centering
    \caption{Reward and automatic reward scaling}
    \label{tab:reward_parameters}
    \begin{tabular}{l|l}
        \hline
        $\alpha_\psi$ & 0.8 \\ \hline
        $\alpha_d$ & 0.01 \\ \hline
        scaling\_learning & 1e-4 \\ \hline
        scaling\_init\_value & 0.5 \\ \hline
        $\hat R$ (auto scaling only) & 0.009 \\ \hline
    \end{tabular}
\end{table}
\begin{table}[h]
    \centering
    \caption{Pretrain Hyperparameters}
    \label{tab:pt_parameters}
    \begin{tabular}{l|l}
        \hline
        learning rate & 1e-3 \\ \hline
        batchsize & 128 \\ \hline
        $N_{pt}$ & 2,976,947 \\ \hline
        num\_epoch & 1 \\ \hline
        $\alpha_{\text{reg}}$ & 1 \\ \hline
        $\alpha_{\psi, \text{pt}}$ & 1e-6 \\ \hline
    \end{tabular}
\end{table}  
\subsection{Additional Reward Experiment Result}
\label{app:amp}
Additional experiment results of the grid search for Exp B and Exp C are listed in Table \ref{tab:exp_deep_mimic} and Tab \ref{tab:exp_amps}, respectively.
\begin{table}[tbh]
\centering
\caption{Additional experiment result for Exp B}
\label{tab:exp_deep_mimic}
\begin{tabular}{l|l|l}
\hline
$\boldsymbol{\alpha^s}$ & \textbf{lap time} & \textbf{MRO}\\ \hline
1      &  \textbf{66.443} $\pm$ 0.029        & \textbf{0.4159} $\pm$ 0.0468 \\ \hline
0.8    &  66.673 $\pm$ 0.020        & \textbf{0.4108} $\pm$ 0.0108 \\ \hline
0.6    &  67.530 $\pm$ 0.038        & 0.4984 $\pm$ 0.0690 \\ \hline
0.4    &  66.543 $\pm$ 0.038        & 0.6428 $\pm$ 0.0296 \\ \hline
0.2    &  66.560 $\pm$ 0.028        & 0.7093 $\pm$ 0.0401 \\ \hline
\end{tabular}
\end{table}

\begin{table}[tbh]
\centering
\caption{Additional experiment result for Exp C}
\label{tab:exp_amps}
\begin{tabular}{l|l|l}
\hline
$\boldsymbol{\alpha^s}$ & \textbf{lap time} & \textbf{MRO}\\ \hline
1      &  $68.588 \pm 0.027$        & $0.1407 \pm 0.0134$ \\ \hline
0.9    &  $68.638 \pm 0.053$        & $0.1519 \pm 0.0097$ \\ \hline
0.7    &  $68.496 \pm 0.066$        & $0.1371 \pm 0.0097$ \\ \hline
0.5    &  $67.960 \pm 0.037$        & $0.1335 \pm 0.0166$ \\ \hline
0.3    &  $67.630 \pm 0.054$        & $\textbf{0.1283} \pm 0.0073$ \\ \hline
0.1    &  $66.796 \pm 0.029$        & $\textbf{0.1240} \pm 0.0058$ \\ \hline
0.09   &  $66.698 \pm 0.030$        & $0.1354 \pm 0.0098$ \\ \hline
0.07   &  $66.608 \pm 0.033$        & $0.1389 \pm 0.0061$ \\ \hline
0.05   &  $\textbf{66.480} \pm 0.020$ & $0.1636 \pm 0.0104$ \\ \hline
0.03   &  $66.547 \pm 0.031$        & $0.1933 \pm 0.0123$ \\ \hline
0.01   &  $\textbf{66.432} \pm 0.060$ & $0.4307 \pm 0.0289$ \\ \hline
\end{tabular}
\end{table}

\subsection{Further evaluations}
\label{app:usecase2}
\begin{figure}[tb]
    \centering
    \includegraphics[width=0.8\columnwidth]{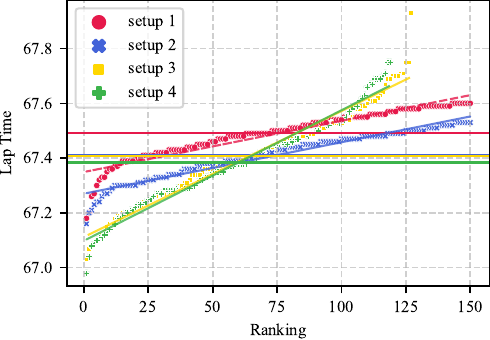}
    \vspace{-8pt}
    \caption{Comparison of four different race car setups by ranking the driven laps by lap time. From Setup 1 to 4, the mechanical balance progressively shifts rearward. The horizontal lines represent the average lap time for each setup.}
    \label{fig:usecase2}
\end{figure}
Except for the use case validation presented in Section \ref{sec:human_eval}, we present an additional analysis conducted on a different track and with a different race car. Before entering the simulator, race engineers aim to determine the optimal mechanical balance setup that provides the best handling characteristics on the track. They propose four setup options (Setup 1–4), with the mechanical balance progressively shifting rearward. Based on their experience, they anticipate that Setup 1 will have controllability issues related to understeer, while Setup 4 may introduce too much oversteer. However, they are uncertain about the optimal balance point between these two extremes.

We follow the same workflow as outlined in Section \ref{sec: usecase} to address this problem. A total of 200 laps are generated, and after filtering outliers and unfinished laps, the remaining data are presented in Fig. \ref{fig:usecase2}, using the same format as Fig. \ref{fig:usecase_ranked_laptime}. Consistent with the race engineers’ prior knowledge, Setup 1 performs the worst in lap time due to the lack of controllability caused by understeer. Setups 3 and 4, which introduce excessive oversteer, show the largest lap time slopes among all four setups, indicating the poorest drivability. Setup 2, which has a similar average lap time to Setups 3 and 4, but with a slope comparable to Setup 1, delivers the best and most consistent performance, demonstrating superior drivability.

\bibliography{main.bib}

@article{deepMimic,
	author = {Peng, Xue Bin and Abbeel, Pieter and Levine, Sergey and van de Panne, Michiel},
	title = {DeepMimic: Example-guided Deep Reinforcement Learning of Physics-based Character Skills},
	journal = {ACM Trans. Graph.},
	issue_date = {August 2018},
	volume = {37},
	number = {4},
	month = jul,
	year = {2018},
	issn = {0730-0301},
	pages = {143:1--143:14},
	articleno = {143},
	numpages = {14},
	url = {http://doi.acm.org/10.1145/3197517.3201311},
	doi = {10.1145/3197517.3201311},
	acmid = {3201311},
	publisher = {ACM},
	address = {New York, NY, USA},
}

@article{self_prediction,
	title = {Understanding {Self}-{Predictive} {Learning} for {Reinforcement} {Learning}},
	url = {http://arxiv.org/abs/2212.03319},
	doi = {10.48550/arXiv.2212.03319},
	urldate = {2025-01-07},
	publisher = {arXiv},
	author = {Tang, Yunhao and Guo, Zhaohan Daniel and Richemond, Pierre Harvey and Pires, Bernardo Ávila and Chandak, Yash and Munos, Rémi and Rowland, Mark and Azar, Mohammad Gheshlaghi and Lan, Charline Le and Lyle, Clare and György, András and Thakoor, Shantanu and Dabney, Will and Piot, Bilal and Calandriello, Daniele and Valko, Michal},
	month = dec,
	year = {2022},
	note = {arXiv:2212.03319 [cs]},
}

@article{amp,
	title = {{AMP}: {Adversarial} {Motion} {Priors} for {Stylized} {Physics}-{Based} {Character} {Control}},
	volume = {40},
	issn = {0730-0301, 1557-7368},
	shorttitle = {{AMP}},
	url = {http://arxiv.org/abs/2104.02180},
	doi = {10.1145/3450626.3459670},
	number = {4},
	urldate = {2025-01-09},
	journal = {ACM Transactions on Graphics},
	author = {Peng, Xue Bin and Ma, Ze and Abbeel, Pieter and Levine, Sergey and Kanazawa, Angjoo},
	month = aug,
	year = {2021},
	note = {arXiv:2104.02180 [cs]},
	pages = {1--20},
}

@article{il_divergence_minimization,
	title = {A {Divergence} {Minimization} {Perspective} on {Imitation} {Learning} {Methods}},
	url = {http://arxiv.org/abs/1911.02256},
	doi = {10.48550/arXiv.1911.02256},
	urldate = {2025-01-09},
	publisher = {arXiv},
	author = {Ghasemipour, Seyed Kamyar Seyed and Zemel, Richard and Gu, Shixiang},
	month = nov,
	year = {2019},
	note = {arXiv:1911.02256 [cs]},
}

@misc{il_divergence_minimization_2,
	title = {Imitation {Learning} as \$f\$-{Divergence} {Minimization}},
	publisher = {arXiv},
	author = {Ke, Liyiming and Choudhury, Sanjiban and Barnes, Matt and Sun, Wen and Lee, Gilwoo and Srinivasa, Siddhartha},
	month = may,
	year = {2020},
	note = {arXiv:1905.12888},
}

@article{bezier_curve,
	title = {Introducing {Probabilistic} {Bézier} {Curves} for {N}-{Step} {Sequence} {Prediction}},
	volume = {34},
	copyright = {https://www.aaai.org},
	issn = {2374-3468, 2159-5399},
	url = {https://ojs.aaai.org/index.php/AAAI/article/view/6576},
	doi = {10.1609/aaai.v34i06.6576},
	number = {06},
	urldate = {2025-01-12},
	journal = {Proceedings of the AAAI Conference on Artificial Intelligence},
	author = {Hug, Ronny and Hübner, Wolfgang and Arens, Michael},
	month = apr,
	year = {2020},
	pages = {10162--10169},
}

@misc{self_predictive_efficient,
	title = {Data-{Efficient} {Reinforcement} {Learning} with {Self}-{Predictive} {Representations}},
	url = {http://arxiv.org/abs/2007.05929},
	doi = {10.48550/arXiv.2007.05929},
	urldate = {2025-01-08},
	publisher = {arXiv},
	author = {Schwarzer, Max and Anand, Ankesh and Goel, Rishab and Hjelm, R. Devon and Courville, Aaron and Bachman, Philip},
	month = may,
	year = {2021},
	note = {arXiv:2007.05929 [cs]},
}

@misc{understanding_self_predictive,
	title = {Bridging {State} and {History} {Representations}: {Understanding} {Self}-{Predictive} {RL}},
	shorttitle = {Bridging {State} and {History} {Representations}},
	url = {http://arxiv.org/abs/2401.08898},
	doi = {10.48550/arXiv.2401.08898},
	urldate = {2025-01-08},
	publisher = {arXiv},
	author = {Ni, Tianwei and Eysenbach, Benjamin and Seyedsalehi, Erfan and Ma, Michel and Gehring, Clement and Mahajan, Aditya and Bacon, Pierre-Luc},
	month = apr,
	year = {2024},
	note = {arXiv:2401.08898 [cs]},
}

@misc{when_self_predict,
	title = {When does {Self}-{Prediction} help? {Understanding} {Auxiliary} {Tasks} in {Reinforcement} {Learning}},
	shorttitle = {When does {Self}-{Prediction} help?},
	url = {http://arxiv.org/abs/2406.17718},
	doi = {10.48550/arXiv.2406.17718},
	urldate = {2025-01-08},
	publisher = {arXiv},
	author = {Voelcker, Claas and Kastner, Tyler and Gilitschenski, Igor and Farahmand, Amir-massoud},
	month = jun,
	year = {2024},
	note = {arXiv:2406.17718 [cs]},
}

@article{world_models,
	title = {World {Models}},
	url = {http://arxiv.org/abs/1803.10122},
	doi = {10.5281/zenodo.1207631},
	urldate = {2025-01-13},
	author = {Ha, David and Schmidhuber, Jürgen},
	month = mar,
	year = {2018},
	note = {arXiv:1803.10122 [cs]},
}

@misc{dreamer,
	title = {Dream to {Control}: {Learning} {Behaviors} by {Latent} {Imagination}},
	shorttitle = {Dream to {Control}},
	url = {http://arxiv.org/abs/1912.01603},
	doi = {10.48550/arXiv.1912.01603},
	urldate = {2025-01-13},
	publisher = {arXiv},
	author = {Hafner, Danijar and Lillicrap, Timothy and Ba, Jimmy and Norouzi, Mohammad},
	month = mar,
	year = {2020},
	note = {arXiv:1912.01603 [cs]},
}

@misc{aux_pointgoal,
	title = {Auxiliary {Tasks} {Speed} {Up} {Learning} {PointGoal} {Navigation}},
	url = {http://arxiv.org/abs/2007.04561},
	doi = {10.48550/arXiv.2007.04561},
	urldate = {2025-01-09},
	publisher = {arXiv},
	author = {Ye, Joel and Batra, Dhruv and Wijmans, Erik and Das, Abhishek},
	month = nov,
	year = {2020},
	note = {arXiv:2007.04561 [cs]},
}

@misc{PPO,
	title = {Proximal {Policy} {Optimization} {Algorithms}},
	publisher = {arXiv},
	author = {Schulman, John and Wolski, Filip and Dhariwal, Prafulla and Radford, Alec and Klimov, Oleg},
	month = aug,
	year = {2017},
	note = {arXiv:1707.06347},
}

@inproceedings{BC,
  title={A Framework for Behavioural Cloning},
  author={Michael Bain and Claude Sammut},
  booktitle={Machine Intelligence 15},
  year={1995},
  url={https://api.semanticscholar.org/CorpusID:10738655}
}

@misc{GAIL,
	title = {Generative {Adversarial} {Imitation} {Learning}},
	url = {http://arxiv.org/abs/1606.03476},
	doi = {10.48550/arXiv.1606.03476},
	urldate = {2025-01-13},
	publisher = {arXiv},
	author = {Ho, Jonathan and Ermon, Stefano},
	month = jun,
	year = {2016},
	note = {arXiv:1606.03476 [cs]},
}

@article{DAgger,
	title = {A {Reduction} of {Imitation} {Learning} and {Structured} {Prediction} to {No}-{Regret} {Online} {Learning}},
	url = {http://arxiv.org/abs/1011.0686},
	doi = {10.48550/arXiv.1011.0686},
	urldate = {2025-01-13},
	publisher = {arXiv},
	author = {Ross, Stephane and Gordon, Geoffrey J. and Bagnell, J. Andrew},
	month = mar,
	year = {2011},
	note = {arXiv:1011.0686 [cs]},
}

@misc{MultiGAIL,
	title = {Generating {Personas} for {Games} with {Multimodal} {Adversarial} {Imitation} {Learning}},
	url = {http://arxiv.org/abs/2308.07598},
	doi = {10.48550/arXiv.2308.07598},
	urldate = {2025-01-13},
	publisher = {arXiv},
	author = {Ahlberg, William and Sestini, Alessandro and Tollmar, Konrad and Gisslén, Linus},
	month = aug,
	year = {2023},
	note = {arXiv:2308.07598 [cs]},
}

@misc{BeTAIL,
	title = {{BeTAIL}: {Behavior} {Transformer} {Adversarial} {Imitation} {Learning} from {Human} {Racing} {Gameplay}},
	shorttitle = {{BeTAIL}},
	url = {http://arxiv.org/abs/2402.14194},
	doi = {10.48550/arXiv.2402.14194},
	urldate = {2025-01-13},
	publisher = {arXiv},
	author = {Weaver, Catherine and Tang, Chen and Hao, Ce and Kawamoto, Kenta and Tomizuka, Masayoshi and Zhan, Wei},
	month = jul,
	year = {2024},
	note = {arXiv:2402.14194 [cs]},
}

@article{modeling_driving_AIL,
	title = {Modeling {Human} {Driving} {Behavior} through {Generative} {Adversarial} {Imitation} {Learning}},
	volume = {24},
	issn = {1524-9050, 1558-0016},
	url = {http://arxiv.org/abs/2006.06412},
	doi = {10.1109/TITS.2022.3227738},
	number = {3},
	urldate = {2025-01-13},
	journal = {IEEE Transactions on Intelligent Transportation Systems},
	author = {Bhattacharyya, Raunak and Wulfe, Blake and Phillips, Derek and Kuefler, Alex and Morton, Jeremy and Senanayake, Ransalu and Kochenderfer, Mykel},
	month = mar,
	year = {2023},
	note = {arXiv:2006.06412 [cs]},
	pages = {2874--2887},
}

@inproceedings{covariate_shift,
	title = {Efficient {Reductions} for {Imitation} {Learning}},
	url = {https://proceedings.mlr.press/v9/ross10a.html},
	language = {en},
	urldate = {2025-01-13},
	booktitle = {Proceedings of the {Thirteenth} {International} {Conference} on {Artificial} {Intelligence} and {Statistics}},
	publisher = {JMLR Workshop and Conference Proceedings},
	author = {Ross, Stephane and Bagnell, Drew},
	month = mar,
	year = {2010},
	note = {ISSN: 1938-7228},
	pages = {661--668},
}

@article{ju_digital_twin_2023,
	title = {Digital {Twin} of a {Driver}-in-the-{Loop} {Race} {Car} {Simulation} {With} {Contextual} {Reinforcement} {Learning}},
	volume = {8},
	copyright = {https://ieeexplore.ieee.org/Xplorehelp/downloads/license-information/IEEE.html},
	issn = {2377-3766, 2377-3774},
	url = {https://ieeexplore.ieee.org/document/10132573/},
	doi = {10.1109/LRA.2023.3279618},
	number = {7},
	urldate = {2025-01-13},
	journal = {IEEE Robotics and Automation Letters},
	author = {Ju, Siwei and Van Vliet, Peter and Arenz, Oleg and Peters, Jan},
	month = jul,
	year = {2023},
	pages = {4107--4114},
}

@article{sonyAI_2022,
	title = {Outracing champion {Gran} {Turismo} drivers with deep reinforcement learning},
	volume = {602},
	issn = {0028-0836, 1476-4687},
	url = {https://www.nature.com/articles/s41586-021-04357-7},
	doi = {10.1038/s41586-021-04357-7},
	language = {en},
	number = {7896},
	urldate = {2024-05-23},
	journal = {Nature},
	author = {Wurman, Peter R. and Barrett, Samuel and Kawamoto, Kenta and MacGlashan, James and Subramanian, Kaushik and Walsh, Thomas J. and Capobianco, Roberto and Devlic, Alisa and Eckert, Franziska and Fuchs, Florian and Gilpin, Leilani and Khandelwal, Piyush and Kompella, Varun and Lin, HaoChih and MacAlpine, Patrick and Oller, Declan and Seno, Takuma and Sherstan, Craig and Thomure, Michael D. and Aghabozorgi, Houmehr and Barrett, Leon and Douglas, Rory and Whitehead, Dion and Dürr, Peter and Stone, Peter and Spranger, Michael and Kitano, Hiroaki},
	month = feb,
	year = {2022},
	pages = {223--228},
}

@inproceedings{benchmark_IL_racing,
	title = {A {Benchmark} {Comparison} of {Imitation} {Learning}-based {Control} {Policies} for {Autonomous} {Racing}},
	url = {https://ieeexplore.ieee.org/abstract/document/10186780},
	doi = {10.1109/IV55152.2023.10186780},
	urldate = {2025-01-08},
	booktitle = {2023 {IEEE} {Intelligent} {Vehicles} {Symposium} ({IV})},
	author = {Sun, Xiatao and Zhou, Mingyan and Zhuang, Zhijun and Yang, Shuo and Betz, Johannes and Mangharam, Rahul},
	month = jun,
	year = {2023},
	note = {ISSN: 2642-7214},
	pages = {1--5},
}

@misc{deepracing,
	title = {{DeepRacing}: {Parameterized} {Trajectories} for {Autonomous} {Racing}},
	shorttitle = {{DeepRacing}},
	url = {http://arxiv.org/abs/2005.05178},
	doi = {10.48550/arXiv.2005.05178},
	urldate = {2025-01-13},
	publisher = {arXiv},
	author = {Weiss, Trent and Behl, Madhur},
	month = may,
	year = {2020},
	note = {arXiv:2005.05178 [cs]},
}

@inproceedings{rl_path_tracking,
	title = {A {Study} of {Reinforcement} {Learning} {Techniques} for {Path} {Tracking} in {Autonomous} {Vehicles}},
	url = {https://ieeexplore.ieee.org/document/10588521},
	doi = {10.1109/IV55156.2024.10588521},
	urldate = {2025-01-13},
	booktitle = {2024 {IEEE} {Intelligent} {Vehicles} {Symposium} ({IV})},
	author = {Chemin, Jason and Hill, Ashley and Lucet, Eric and Mayoue, Aurélien},
	month = jun,
	year = {2024},
	note = {ISSN: 2642-7214},
	pages = {1442--1449},
}

@misc{mega_dagger,
	title = {{MEGA}-{DAgger}: {Imitation} {Learning} with {Multiple} {Imperfect} {Experts}},
	shorttitle = {{MEGA}-{DAgger}},
	url = {http://arxiv.org/abs/2303.00638},
	doi = {10.48550/arXiv.2303.00638},
	urldate = {2025-01-13},
	publisher = {arXiv},
	author = {Sun, Xiatao and Yang, Shuo and Zhou, Mingyan and Liu, Kunpeng and Mangharam, Rahul},
	month = may,
	year = {2024},
	note = {arXiv:2303.00638 [cs]},
}

@article{autonomous_on_edge,
	title = {Autonomous {Vehicles} on the {Edge}: {A} {Survey} on {Autonomous} {Vehicle} {Racing}},
	volume = {3},
	issn = {2687-7813},
	shorttitle = {Autonomous {Vehicles} on the {Edge}},
	url = {http://arxiv.org/abs/2202.07008},
	doi = {10.1109/ojits.2022.3181510},
	urldate = {2025-01-13},
	journal = {IEEE Open Journal of Intelligent Transportation Systems},
	author = {Betz, Johannes and Zheng, Hongrui and Liniger, Alexander and Rosolia, Ugo and Karle, Phillip and Behl, Madhur and Krovi, Venkat and Mangharam, Rahul},
	year = {2022},
	note = {arXiv:2202.07008 [cs]},
	pages = {458--488},
}

@inproceedings{driving_fuzzy,
	title = {Driving style imitation in simulated car racing using style evaluators and multi-objective evolution of a fuzzy logic controller},
	url = {https://ieeexplore.ieee.org/document/6893872},
	doi = {10.1109/NORBERT.2014.6893872},
	urldate = {2025-01-13},
	booktitle = {2014 {IEEE} {Conference} on {Norbert} {Wiener} in the 21st {Century} ({21CW})},
	author = {Wang, Tsaipei and Liaw, Keng-Te},
	month = jun,
	year = {2014},
	pages = {1--7},
}

@Book{milliken:1995:RCVD,
  author    = {Milliken, William F. and Milliken, Douglas L.},
  publisher = {Society of Automotive Engineers Warrendale},
  title     = {Race Car Vehicle Dynamics},
  year      = {1995},
}

@inproceedings{pure_pursuit,
	title = {Implementation of the {Pure} {Pursuit} {Path} {Tracking} {Algorithm}},
	url = {https://www.semanticscholar.org/paper/Implementation-of-the-Pure-Pursuit-Path-Tracking-Coulter/ee756e53b6a68cb2e7a2e5d537a3eff43d793d70},
	urldate = {2025-01-13},
	author = {Coulter, Craig},
	year = {1992},
}

@inproceedings{stanley,
	title = {Autonomous {Automobile} {Trajectory} {Tracking} for {Off}-{Road} {Driving}: {Controller} {Design}, {Experimental} {Validation} and {Racing}},
	shorttitle = {Autonomous {Automobile} {Trajectory} {Tracking} for {Off}-{Road} {Driving}},
	doi = {10.1109/ACC.2007.4282788},
	author = {Hoffmann, Gabriel and Tomlin, Claire and Montemerlo, Michael and Thrun, Sebastian},
	month = aug,
	year = {2007},
	pages = {2296--2301},
}

@article{mpc_survey,
	title = {Review and {Comparison} of {Path} {Tracking} {Based} on {Model} {Predictive} {Control}},
	volume = {8},
	copyright = {http://creativecommons.org/licenses/by/3.0/},
	issn = {2079-9292},
	url = {https://www.mdpi.com/2079-9292/8/10/1077},
	doi = {10.3390/electronics8101077},
	language = {en},
	number = {10},
	urldate = {2025-01-13},
	journal = {Electronics},
	author = {Bai, Guoxing and Meng, Yu and Liu, Li and Luo, Weidong and Gu, Qing and Liu, Li},
	month = oct,
	year = {2019},
	note = {Number: 10
Publisher: Multidisciplinary Digital Publishing Institute},
	pages = {1077},
}

@article{lockel_adaptive_2023,
	title = {An {Adaptive} {Human} {Driver} {Model} for {Realistic} {Race} {Car} {Simulations}},
	volume = {53},
	copyright = {https://ieeexplore.ieee.org/Xplorehelp/downloads/license-information/IEEE.html},
	issn = {2168-2216, 2168-2232},
	url = {https://ieeexplore.ieee.org/document/10173774/},
	doi = {10.1109/TSMC.2023.3285588},
	number = {11},
	urldate = {2025-01-13},
	journal = {IEEE Transactions on Systems, Man, and Cybernetics: Systems},
	author = {Löckel, Stefan and Ju, Siwei and Schaller, Maximilian and Van Vliet, Peter and Peters, Jan},
	month = nov,
	year = {2023},
	pages = {6718--6730},
}

@article{lockel_identification_2022,
	title = {Identification and modelling of race driving styles},
	volume = {60},
	issn = {0042-3114, 1744-5159},
	url = {https://www.tandfonline.com/doi/full/10.1080/00423114.2021.1930070},
	doi = {10.1080/00423114.2021.1930070},
	language = {en},
	number = {8},
	urldate = {2025-01-14},
	journal = {Vehicle System Dynamics},
	author = {Löckel, Stefan and Kretschi, André and Van Vliet, Peter and Peters, Jan},
	month = aug,
	year = {2022},
	pages = {2890--2918},
}

@inproceedings{robust_player_imitation,
	title = {Robust player imitation using multiobjective evolution},
	url = {https://ieeexplore.ieee.org/document/4983007},
	doi = {10.1109/CEC.2009.4983007},
	urldate = {2025-01-15},
	booktitle = {2009 {IEEE} {Congress} on {Evolutionary} {Computation}},
	author = {van Hoorn, Niels and Togelius, Julian and Wierstra, Daan and Schmidhuber, Jurgen},
	month = may,
	year = {2009},
	note = {ISSN: 1941-0026},
	pages = {652--659},
}

@misc{CAP_survey,
	title = {A {Survey} of {Temporal} {Credit} {Assignment} in {Deep} {Reinforcement} {Learning}},
	url = {http://arxiv.org/abs/2312.01072},
	doi = {10.48550/arXiv.2312.01072},
	urldate = {2025-01-21},
	publisher = {arXiv},
	author = {Pignatelli, Eduardo and Ferret, Johan and Geist, Matthieu and Mesnard, Thomas and Hasselt, Hado van and Pietquin, Olivier and Toni, Laura},
	month = jul,
	year = {2024},
	note = {arXiv:2312.01072 [cs]},
}

@inproceedings{dynamic_utility,
	address = {New York, NY, USA},
	series = {{ICML} '05},
	title = {Dynamic preferences in multi-criteria reinforcement learning},
	isbn = {978-1-59593-180-1},
	url = {https://dl.acm.org/doi/10.1145/1102351.1102427},
	doi = {10.1145/1102351.1102427},
	urldate = {2025-01-22},
	booktitle = {Proceedings of the 22nd international conference on {Machine} learning},
	publisher = {Association for Computing Machinery},
	author = {Natarajan, Sriraam and Tadepalli, Prasad},
	month = aug,
	year = {2005},
	pages = {601--608},
}

@article{multi_objective_survey,
	title = {A {Practical} {Guide} to {Multi}-{Objective} {Reinforcement} {Learning} and {Planning}},
	volume = {36},
	issn = {1387-2532, 1573-7454},
	url = {http://arxiv.org/abs/2103.09568},
	doi = {10.1007/s10458-022-09552-y},
	number = {1},
	urldate = {2025-01-22},
	journal = {Autonomous Agents and Multi-Agent Systems},
	author = {Hayes, Conor F. and Rădulescu, Roxana and Bargiacchi, Eugenio and Källström, Johan and Macfarlane, Matthew and Reymond, Mathieu and Verstraeten, Timothy and Zintgraf, Luisa M. and Dazeley, Richard and Heintz, Fredrik and Howley, Enda and Irissappane, Athirai A. and Mannion, Patrick and Nowé, Ann and Ramos, Gabriel and Restelli, Marcello and Vamplew, Peter and Roijers, Diederik M.},
	month = apr,
	year = {2022},
	note = {arXiv:2103.09568 [cs]},
	pages = {26},
}

@article{multi_objective_survey_2,
	title = {A {Survey} of {Multi}-{Objective} {Sequential} {Decision}-{Making}},
	volume = {48},
	issn = {1076-9757},
	url = {http://arxiv.org/abs/1402.0590},
	doi = {10.1613/jair.3987},
	urldate = {2025-01-22},
	journal = {Journal of Artificial Intelligence Research},
	author = {Roijers, Diederik Marijn and Vamplew, Peter and Whiteson, Shimon and Dazeley, Richard},
	month = oct,
	year = {2013},
	note = {arXiv:1402.0590 [cs]},
	pages = {67--113},
}

@article{mo_utility_survey,
	title = {Multi-{Objective} {Multi}-{Agent} {Decision} {Making}: {A} {Utility}-based {Analysis} and {Survey}},
	volume = {34},
	issn = {1387-2532, 1573-7454},
	shorttitle = {Multi-{Objective} {Multi}-{Agent} {Decision} {Making}},
	url = {http://arxiv.org/abs/1909.02964},
	doi = {10.1007/s10458-019-09433-x},
	number = {1},
	urldate = {2025-01-22},
	journal = {Autonomous Agents and Multi-Agent Systems},
	author = {Rădulescu, Roxana and Mannion, Patrick and Roijers, Diederik M. and Nowé, Ann},
	month = apr,
	year = {2020},
	note = {arXiv:1909.02964 [cs]},
	pages = {10},
}

@InProceedings{promp,
  author       = {Paraschos, Alexandros and Rueckert, Elmar and Peters, Jan and Neumann, Gerhard},
  booktitle    = {IEEE/RSJ International Conference on Intelligent Robots and Systems (IROS)},
  title        = {Model-Free Probabilistic Movement Primitives for Physical Interaction},
  year         = {2015},
  organization = {IEEE},
  pages        = {2860--2866},
}

@article{contextual_rl_2015,
	title = {Contextual {Markov} {Decision} {Processes}},
	url = {http://arxiv.org/abs/1502.02259},
	doi = {10.48550/arXiv.1502.02259},
	urldate = {2025-01-31},
	publisher = {arXiv},
	author = {Hallak, Assaf and Castro, Dotan Di and Mannor, Shie},
	month = feb,
	year = {2015},
	note = {arXiv:1502.02259 [stat]},
}

@article{sbrl,
    author = {Ju, Siwei and Arenz, Oleg and van Vliet, Peter and Peters, Jan},
    title = {Style-biased reinforcement learning for quadruped locomotion},
    journal = {Under review at ICRA},
    year ={2025}, 
}

@article{optimal_control,
author = {Nicola Dal Bianco and Enrico Bertolazzi and Francesco Biral and Matteo Massaro},
title = {Comparison of direct and indirect methods for minimum lap time optimal control problems},
journal = {Vehicle System Dynamics},
volume = {57},
number = {5},
pages = {665--696},
year = {2019},
publisher = {Taylor \& Francis},
doi = {10.1080/00423114.2018.1480048},


URL = { 
    
        https://doi.org/10.1080/00423114.2018.1480048
    
    

},
eprint = { 
    
        https://doi.org/10.1080/00423114.2018.1480048
    
    

}

}

@article{mpc,
	title = {Learning-{Based} {Model} {Predictive} {Control} for {Autonomous} {Racing}},
	volume = {4},
	copyright = {https://ieeexplore.ieee.org/Xplorehelp/downloads/license-information/IEEE.html},
	issn = {2377-3766, 2377-3774},
	url = {https://ieeexplore.ieee.org/document/8754713/},
	doi = {10.1109/LRA.2019.2926677},
	number = {4},
	urldate = {2025-10-17},
	journal = {IEEE Robotics and Automation Letters},
	author = {Kabzan, Juraj and Hewing, Lukas and Liniger, Alexander and Zeilinger, Melanie N.},
	month = oct,
	year = {2019},
	pages = {3363--3370},
}
\bibliographystyle{plainnat}

\end{document}